\documentclass[11pt]{article}

\usepackage{acl}

\usepackage{times}
\usepackage{latexsym}
\usepackage[T1]{fontenc}
\usepackage[utf8]{inputenc}
\usepackage{microtype}
\usepackage{inconsolata}

\usepackage{amsmath,amsfonts,bm}

\def\eqref#1{equation~\ref{#1}}

\def\1{\bm{1}}

\DeclareMathAlphabet{\mathsfit}{\encodingdefault}{\sfdefault}{m}{sl}
\SetMathAlphabet{\mathsfit}{bold}{\encodingdefault}{\sfdefault}{bx}{n}

\usepackage{hyperref}
\usepackage{url}

\usepackage{amsmath}
\usepackage{amssymb}
\usepackage{amsfonts}
\usepackage{amsmath}
\usepackage{arydshln}
\usepackage{amsmath}
\usepackage{adjustbox}

\usepackage{graphicx}
\usepackage{booktabs}
\usepackage{multirow}
\usepackage{placeins}
\usepackage{graphicx}

\usepackage[most]{tcolorbox}
\usepackage{listings}
\usepackage{array}
\usepackage{booktabs}
\usepackage{adjustbox}
\usepackage{multirow}
\usepackage{makecell}

\usepackage{tabularx}

\lstdefinestyle{promptstyle}{
    basicstyle=\ttfamily\footnotesize,
    breaklines=true,
    columns=fullflexible,
    keepspaces=true,
    showstringspaces=false,
    aboveskip=0pt,
    belowskip=0pt,
    xleftmargin=0pt,
    frame=none
}

\newtcblisting{promptbox}{
    listing only,
    enhanced,
    breakable,
    colback=gray!2,
    colframe=black!45,
    boxrule=0.45pt,
    arc=1mm,
    left=2mm,
    right=2mm,
    top=1.5mm,
    bottom=1.5mm,
    listing options={
        style=promptstyle
    }
}

\newcommand{\rA}[1]{\colorbox{accA}{\strut #1}}
\newcommand{\rB}[1]{\colorbox{accB}{\strut #1}}
\newcommand{\rC}[1]{\colorbox{accC}{\strut #1}}
\newcommand{\rD}[1]{\colorbox{accD}{\strut #1}}
\newcommand{\rE}[1]{\colorbox{accE}{\strut #1}}

\newcommand{\frontierrow}[5]{%
\noindent
\begin{tabular}{
    @{}
    >{\centering\arraybackslash}m{0.28\linewidth}
    @{\hspace{0.025\linewidth}}
    >{\raggedright\arraybackslash}m{0.67\linewidth}
    @{}
}
\includegraphics[
    width=0.92\linewidth,
    height=3.0cm,
    keepaspectratio
]{#1}
&
\begin{minipage}{\linewidth}
\raggedright

#2
\hfill
\textbf{Profile:} #4

\vspace{0.15em}

{\scriptsize\ttfamily\detokenize{#3}}

\vspace{0.45em}

\textbf{Question.} #5
\end{minipage}
\end{tabular}
}

\usepackage{booktabs}
\usepackage{multirow}
\usepackage[table]{xcolor}
\usepackage{makecell}
\usepackage{array}
\usepackage{adjustbox}

\usepackage{longtable}

\definecolor{accA}{RGB}{198,239,206} 
\definecolor{accB}{RGB}{226,239,218} 
\definecolor{accC}{RGB}{255,242,204} 
\definecolor{accD}{RGB}{248,203,173} 
\definecolor{accE}{RGB}{244,204,204} 
\definecolor{collapseStrong}{RGB}{192,0,0}
\definecolor{collapseRow}{RGB}{252,228,214}

\newcommand{\collapsemark}[1]{%
  \cellcolor{collapseStrong}\textcolor{white}{\textbf{#1}}%
}

\title{ReSI-Math: Exploring Recoverable Capability Frontiers for Efficient Reasoning Improvement
}

\author{
\begin{tabular}{c}
Yao Cui$^{1,*}$ \quad Zhen Yang$^{2,*}$ \quad Yuandong Wang$^{1,\dagger}$ \\
Yangfu Zhu$^{1}$ \quad Tianyu Wo$^{3}$ \quad Zhenzhou Shao$^{1}$ \\[4pt]
$^{1}$Capital Normal University \quad
$^{2}$Tsinghua University \quad
$^{3}$Beihang University \\[2pt]
$^{*}$Equal contribution \quad
$^{\dagger}$Corresponding author
\end{tabular}
}

\begin{document}


\maketitle

\begin{abstract}


Reinforcement learning with verifiable rewards (RLVR) has become an effective approach for improving mathematical reasoning in large language models. However, a fundamental question remains: when a model fails to solve a problem, does it truly lack the underlying capability? In this work, we reveal that direct failure does not necessarily imply inability. Some reasoning failures can be recovered by exploring the model's own trajectories, exposing a \textit{recoverable capability frontier}: a policy-dependent region where solutions are not reliably generated through direct decoding but can be elicited through trajectory-level search. Based on this observation, we introduce \textbf{ReSI-Math}, a framework that explores recoverable capability frontiers and converts recovered capabilities into data-efficient RLVR supervision. ReSI-Math uses Self-Judge-guided evolutionary search to identify recoverable failures under a fixed policy and adaptively selects frontier problems for sequential RLVR as the policy evolves.
On Qwen3-VL-8B, correct--incorrect trajectory crossover produces 272 verifier-confirmed correct offspring from 621 attempts (43.8\%). After three sequential one-shot RLVR stages, the MATH500 and We-Math average improves from 78.71 to 80.16. Under matched single-problem RLVR, random selection reaches only 53.31 versus 54.16 for frontier-guided Q1. Project link:  \url{https://cnu-bot-group.github.io/ReSI-Math/}.

\end{abstract}

\section{Introduction}

\begin{figure*}[t]
    \centering
    \includegraphics[width=\textwidth]{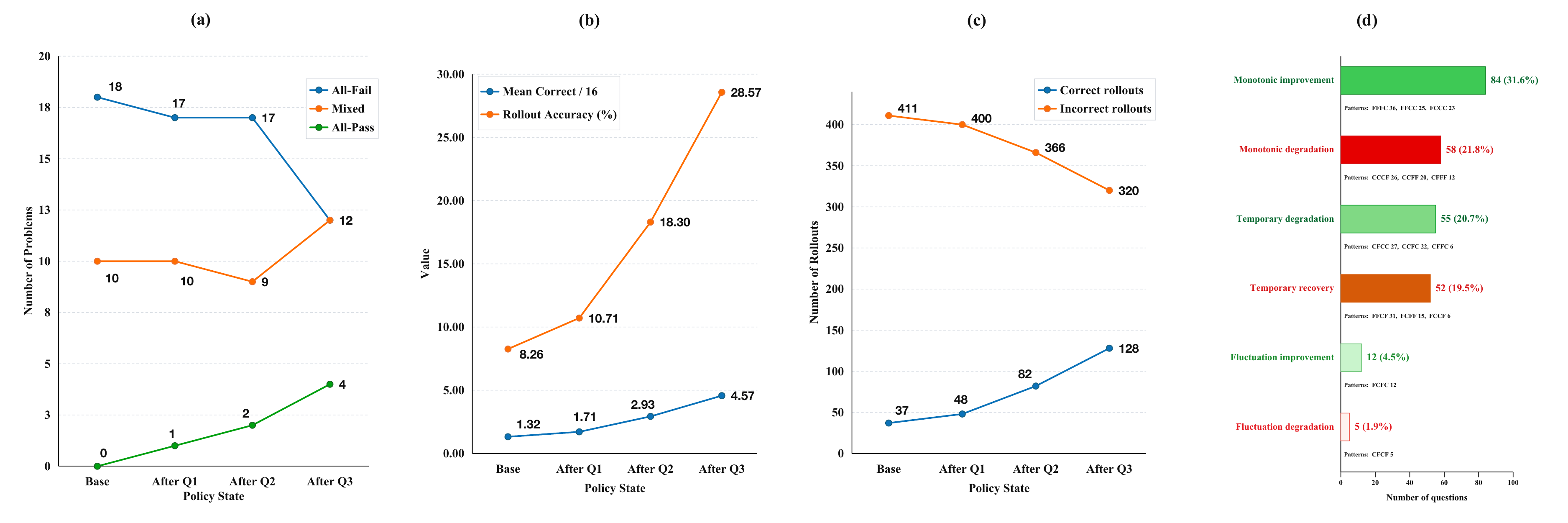}
    \caption{Rollout-profile dynamics over the 28-problem candidate pool and the validation set.
    }
    \label{fig:rollout_profile_dynamics}
\vspace{-2ex}
\end{figure*}

Reinforcement learning with verifiable rewards (RLVR) has become an effective approach to improving mathematical reasoning in large language models, using verifiable outcomes to optimize reasoning policies~\cite{shao2024deepseekmath,guo2025deepseekr1}. Despite these advances, interpreting a model’s failures remains important for understanding and improving its reasoning. An incorrect response leaves open whether the model cannot produce a valid solution or whether the particular generation process failed to elicit one. This raises a fundamental question: \textit{when a model fails to solve a problem, does it truly lack the underlying capability?}

Evidence from inference-time reasoning suggests that a model's observed performance depends on how its reasoning is elicited. Repeated sampling and self-refinement can produce successful solutions after unsuccessful attempts~\cite{cobbe2021Best_of_N,madaan2023selfrefine}, while CoTEvol shows how reasoning trajectories can be evolved through crossover and mutation~\cite{wang2026cotevol}. These findings motivate examining recoverability under fixed model parameters: whether structured exploration of the current model's own trajectories can produce a verifier-confirmed solution when direct generation remains unreliable. This perspective makes the conditions under which reasoning succeeds part of the object of study.

We characterize a \emph{recoverable capability frontier} as a policy-dependent set of problems whose solutions are not reliably generated through direct decoding but can be elicited through bounded trajectory-level search. Direct sampling describes how consistently a problem is solved, while search probes whether a correct solution can be obtained by examining and recombining generated trajectories. Throughout this exploration, the model parameters remain fixed and an external verifier checks the resulting answers. The frontier is thus defined operationally with respect to the current policy and a specified search procedure and budget.

Recoverability also provides a possible connection between capability exploration and subsequent learning. A problem whose solution can be elicited through search, but is not yet reliably generated directly, may offer a useful target for policy optimization. This possibility is particularly relevant to data-efficient RLVR. LIMR~\cite{li2025limr} demonstrates the value of carefully selected training subsets, and One-Shot RLVR~\cite{wang2025oneshotrlvr} shows that even a single problem can support substantial reasoning improvements. These results motivate investigating whether search-confirmed recoverability can help identify problems from which the current policy can learn effectively.

Based on this perspective, we introduce \textbf{ReSI-Math}, a framework that explores recoverable capability frontiers and uses the resulting signals for reasoning improvement. Under a fixed policy, the model first samples multiple reasoning trajectories and then acts as a Self-Judge to analyze their useful components, potential errors, and complementary information. Evolutionary crossover generates new trajectories from these candidates, and a rule-based verifier determines whether the resulting answers are correct. Problems exhibiting successful recovery become candidates for Solver-RLVR, which optimizes fresh rollouts on the original questions using verifiable outcome rewards. The same exploration also produces verifier-labeled trajectories and exposes self-evaluation errors, providing supervision for separately trained Judge-RLVR.

The learning process must also account for changes in the policy. Figure~\ref{fig:rollout_profile_dynamics} shows how observed rollout profiles change across training stages over 28 candidate problems, each evaluated with 16 rollouts. From the base policy to the third stage, the number of all-fail problems decreases from 18 to 12, while the number of all-pass problems increases from 0 to 4; total correct rollouts rise from 37/448 to 128/448. These changes motivate reassessing candidate problems under the updated policy. Building on this motivation, we propose \textit{policy-adaptive sequential one-shot RLVR}. After each training stage, ReSI-Math reprofiles the updated policy and selects the next frontier problem, yielding a local recursive self-improvement loop.

We evaluate ReSI-Math using Qwen3-VL backbones with 2B, 4B, and 8B parameters. On Qwen3-VL-8B, crossover between a correct and an incorrect trajectory produces 272 verifier-confirmed correct offspring from 621 attempts (43.8\%). Three sequential one-shot RLVR stages improve mean accuracy on MATH500 and WE-MATH from 78.71 to 80.16. In a separate single-problem comparison with matched training budgets, frontier-guided selection achieves an average of 54.16 across four mathematical reasoning benchmarks, compared with 53.31 for a random-selection baseline. 



Our contributions are threefold:
\begin{itemize}
    \item We introduce an operational characterization of recoverable capability frontiers, connecting unreliable direct generation with verifier-confirmed recovery through bounded search under fixed model parameters.
    \item We develop ReSI-Math, a Self-Judge-guided evolutionary framework that identifies candidate problems for Solver-RLVR and produces trajectory-level supervision for Judge-RLVR.
    \item We propose policy-adaptive sequential one-shot RLVR and evaluate frontier-guided learning across three model scales, with analyses of search outcomes, training schedules, and changes in problem solvability.
\end{itemize}

\section{Related Work}

Recent studies on reinforcement learning with verifiable rewards (RLVR) increasingly suggest that mathematical reasoning can improve with surprisingly limited training data. LIMR~\citep{li2025limr} studies data selection for RL scaling and shows that carefully selected subsets can retain strong performance with substantially fewer training examples. One-Shot RLVR~\citep{wang2025oneshotrlvr} pushes this regime to the extreme, demonstrating that a single training example can sometimes induce substantial reasoning gains, while DoPR~\citep{zhang2026dopr} dynamically selects informative examples for policy refinement. Related evidence also appears beyond RLVR: One-Shot OPD~\citep{fu2026oneshotopd} shows that a single query can recover a large fraction of full-data on-policy distillation gains. More broadly, data-efficient instruction tuning has shown that carefully selected subsets can outperform or approach much larger training sets~\citep{chen2024alpagasus,xia2024less,ivison2025dataselection}, and the ordering of training examples can itself affect generalization~\citep{he2025dataarrangement}. These results motivate studying not only how much data is required, but also which examples are most valuable. Our work further treats this value as policy-dependent: the most informative problem may change as the policy itself evolves.

\begin{figure*}[t]
    \centering
    \includegraphics[width=\textwidth]{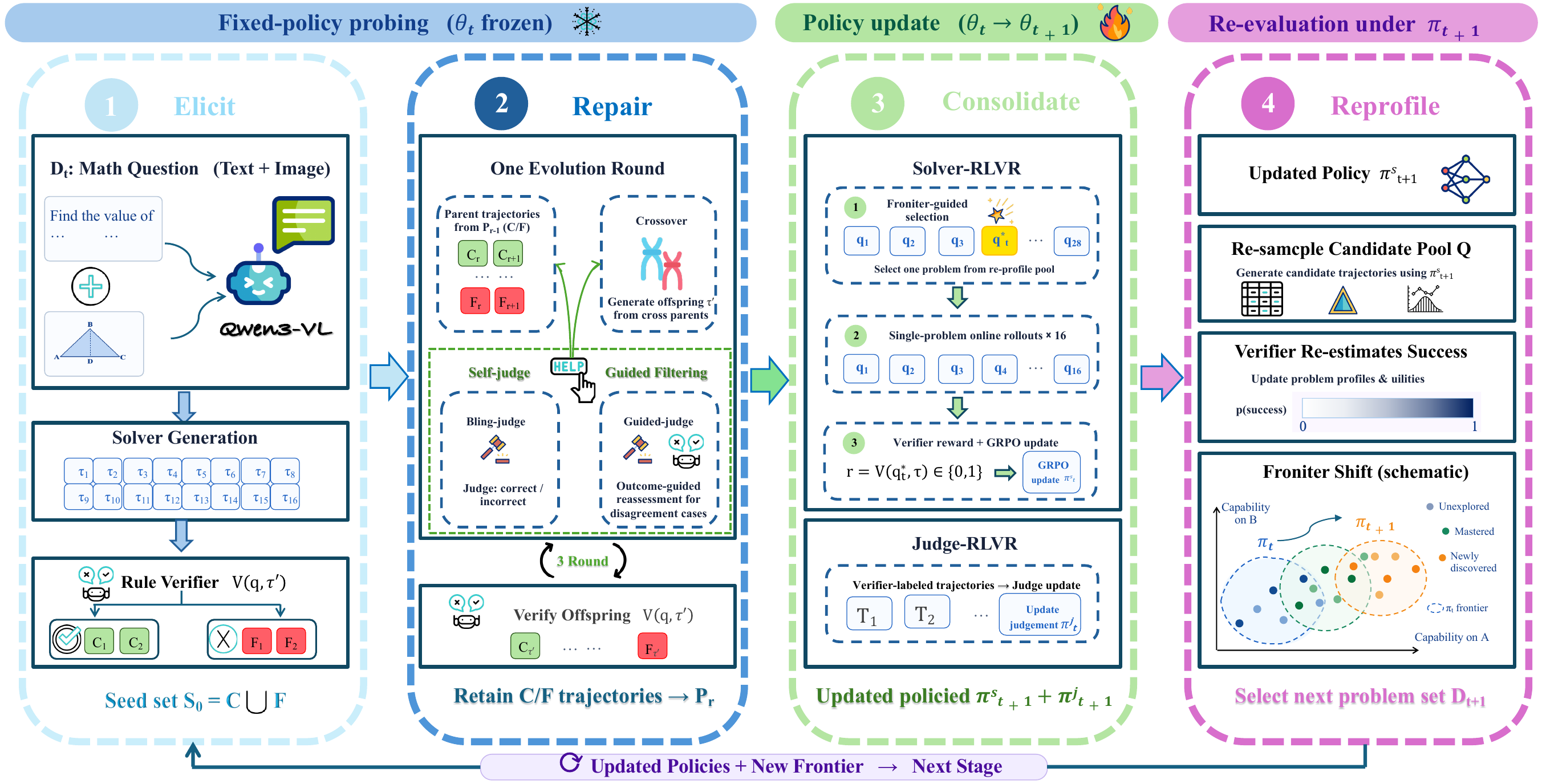}
    \caption{Overview of ReSI-Math, including four stages: Elicit, Repair, Consolidate, and Reprofile.
    }
    \label{fig:ReSI_math_overview}
\end{figure*}

Another line of work improves reasoning through repeated sampling, self-refinement, or evolutionary search. Best-of-N\cite{cobbe2021Best_of_N} and Self-Refine~\citep{madaan2023selfrefine} obtain better solutions through sampling or iterative feedback, while COTEVOL~\citep{wang2026cotevol} formulates CoT synthesis as population-based evolutionary search. In contrast, we use evolution as a \textit{capability-probing mechanism}: rather than focusing only on the final improved CoT, we study whether failed reasoning can be recovered under a fixed policy through trajectory recombination.

Prior work has explored using model-generated feedback to improve subsequent reasoning or generation, such as Self-Refine~\citep{madaan2023selfrefine}. Related LLM-as-a-Judge methods use language models to evaluate model-generated responses~\citep{zheng2023LLM_as_a_judge}. More recent self-evolving systems further organize such feedback into iterative improvement loops; for example, R-Zero~\citep{Huang2026R-Zero} improves Challenger and Solver roles through repeated interaction. ReSI-Math instead realizes a local, policy-adaptive recursion in which Self-Judge guides trajectory search, RLVR updates the policy, and the updated policy is reprofiled to determine the next learning target.





\section{Methodology}
In this section, we introduce \textbf{ReSI-Math}, a framework designed to explore \underline{\textbf{Re}}cursive \underline{\textbf{S}}elf-\underline{\textbf{I}}mprovement on \underline{\textbf{Math}} reasoning problems. 
The core idea of \textbf{ReSI-Math} is that a model's capability boundary need not be observed only after parameter updates through downstream evaluation; instead, it can be actively probed under a fixed policy by conducting structured search over the model's own reasoning trajectories. As illustrated in Figure~\ref{fig:ReSI_math_overview}, ReSI-Math decomposes the resulting self-improvement procedure into four stages: \textbf{Elicit}, \textbf{Repair}, \textbf{Consolidate}, and \textbf{Reprofile}.

\subsection{Elicit: Probing the Recoverable Capability Frontier} 
Given a current policy $\pi_\theta$ and a mathematical problem $x$, the set of sampled reasoning trajectories is denoted as: $\mathcal{T}(x;\pi_\theta)=\{\tau_1,\ldots,\tau_K\}, \tau_i \sim \pi_\theta(\cdot \mid x)$. 
A binary outcome $V(x,\tau_i)\in\{0,1\}$ is assigned to each trajectory by a rule-based verifier $V$. Verifier-confirmed correct and failed trajectories are collected into $C$ and $F$, respectively, forming the initial seed set for subsequent repair, $S_0=C\cup F$. 


The resulting trajectory partition characterizes the direct-generation behavior of the current policy on $x$. Directly expressed capability is characterized by verifier-confirmed success under direct generation, whereas recoverable capability is evaluated with the model parameters $\theta$ held fixed and is defined by the existence of a bounded search or recombination operator $g$ such that 
\begin{equation} \exists\,\tau_i,\tau_j\in\mathcal{T}(x;\pi_\theta), \text{s.t.} V\!\left(x,g(\tau_i,\tau_j)\right)=1. 
\end{equation} 
A successful recovery indicates capability present in the trajectory distribution induced by the current policy rather than capability acquired through parameter updates. The problem set satisfying this criterion is referred as the policy's \textit{recoverable capability frontier}, denoted by $\mathcal{F}_{\mathrm{rec}}(\pi_\theta)$. 

This frontier is not equivalent to a simple correct--incorrect boundary: direct sampling may yield both correct and failed trajectories, while even two failed parent trajectories may be recombined into a verifier-confirmed correct offspring. Such $F\!-\!F\!\rightarrow C$ (Failed--Failed-to-Correct) transitions show that recoverability does not require an initially correct parent. Accordingly, the frontier is not characterized solely by the initial pass rate; the existence of a failure-to-success transition under bounded structured search serves as the operational criterion for recoverability. Under this formulation, \textbf{Elicit} uses direct-generation profiles to localize candidate frontier regions, while recoverability is established in \textbf{Repair} through verifier-confirmed failure-to-success transitions.

\subsection{Repair: Self-Judge-Guided Evolutionary Search} \label{sec:repair} 

Given candidate regions near the recoverable capability frontier, \textbf{Repair} evaluates whether failed reasoning can be transformed into verifier-confirmed correct solutions through Self-Judge-guided evolutionary search. The model parameters remain fixed throughout this stage, and search is restricted to the trajectory space induced by the current policy. 

Self-Judge provides structured analysis of candidate trajectories for evolutionary search. Two complementary judgment modes are considered: the \textbf{Blind Judge} evaluates reasoning without access to verifier outcomes or reference answers, whereas the \textbf{Guided Judge} additionally receives verifier-confirmed outcome feedback while remaining isolated from the reference solution. Self-Judge and the verifier therefore serve distinct functions: Self-Judge provides search guidance, whereas the verifier supplies the external correctness criterion used to determine whether repair succeeds. 

Given parent $\tau_i$, $\tau_j$, and their Self-Judge feedback, an offspring trajectory is generated: \begin{equation} \tau_{\mathrm{child}} \sim \pi_\theta\!\left(\cdot \mid x,\tau_i,\tau_j,J(\tau_i),J(\tau_j)\right). \end{equation} The crossover operator is adapted from the reflective global crossover formulation of CoTEvol~\cite{wang2026cotevol} and is used to recombine complementary reasoning structures rather than reproduce a parent trajectory verbatim. 

Each offspring is subsequently evaluated by the rule-based verifier. A verifier-confirmed transition from at least one failed parent to a correct offspring constitutes a realized failure-to-success transition, including both $C\!-\!F\!\rightarrow C$ and $F\!-\!F\!\rightarrow C$. Such transitions provide operational evidence of recoverability under the fixed policy.
The effectiveness of $C$--$F$ repair is quantified by the offspring-level recovery rate:
\begin{equation}
\mathrm{RRate}_{C\text{-}F}
=
\frac{\#\{\text{correct $C$--$F$ offspring}\}}
{\#\{\text{$C$--$F$ offspring}\}}.
\end{equation}
This metric measures the fraction of generated $C$--$F$ offspring that are verified as correct and is distinct from the fraction of source problems for which at least one successful repair is obtained.
Detailed population construction, Self-Judge protocols, evolutionary updates, and search configurations are provided in Appendix~\ref{app:evolution}, while the crossover prompt is provided in Appendix~\ref{app:crossover_prompt}.

\subsection{Consolidate: Dual-Signal RLVR}\textbf{}

The Repair stage reveals capabilities that can be recovered through search under a fixed policy, but search-time recoverability does not imply that such capabilities are stably expressed by the model. The Consolidate stage therefore converts the signals exposed during evolution into parameter-level learning through RLVR.

\paragraph{Frontier-Guided Problem Selection.} Before Solver-RLVR, a single training problem is selected from the current candidate pool using a policy-dependent top-1 selector.

For candidate problem $q_i$, $K$ denotes the number of sampled trajectories, $N_i^{\mathrm{correct}}$ denotes the number of verifier-confirmed correct trajectories, and $N_i^{\mathrm{parsed}}$ denotes the number of successfully parsed outputs. The rollout success rate is defined as $p_i=N_i^{\mathrm{correct}}/K$. 
Answer-level disagreement among successfully parsed outputs is defined as
\begin{equation}
D_{\mathrm{semantic},i}=1-\sum_k q_{ik}^{2}.
\end{equation}
Here, $q_{ik}=n_{ik}/N_i^{\mathrm{parsed}}$ denotes the proportion of successfully parsed outputs of problem $q_i$ that belong to answer category $k$, where $n_{ik}$ is the number of outputs assigned to category $k$.
The valid-output rate is defined as
\begin{equation}
\mathrm{ParseRate}_i=\frac{N_i^{\mathrm{parsed}}}{K}.
\end{equation}
The frontier-selection score is defined as
\begin{equation}
S_i=p_i(1-p_i)\cdot D_{\mathrm{semantic},i}\cdot \mathrm{ParseRate}_i.
\end{equation}
The highest-scoring candidate is selected as
\begin{equation}
q^\star=\arg\max_{q_i\in\mathcal{Q}} S_i.
\end{equation}

The three factors respectively favor candidates with non-degenerate rollout outcomes, high answer-level disagreement, and reliable output parsing. The selected problem $q^\star$ is then used as the training target for single-problem Solver-RLVR.

\paragraph{Solver Consolidation.}
For the Solver, a problem that exhibits a verifier-confirmed failure-to-success transition is treated as a recoverable training example. Rather than imitating the evolved correct trajectory itself, Solver-RLVR trains the current policy directly on the original problem using verifier-based rewards. The objective is to transform capability that can only be elicited through search into behavior that can be expressed more reliably through direct generation.

\paragraph{Judge Consolidation.}
Evolution also exposes self-evaluation errors by comparing Self-Judge predictions with verifier-confirmed outcomes. These disagreements provide trajectory-level supervision for Judge-RLVR, where the model is trained to better discriminate whether a candidate reasoning trajectory reaches a correct outcome.

The two learning signals are therefore complementary: recoverable problems provide supervision for Solver-RLVR, while judgment errors provide trajectory-level supervision for Judge-RLVR.

Thus, Consolidate converts search-time behavior into parameterized capability: recoverable reasoning becomes more stable direct problem-solving behavior, while self-evaluation errors become supervision for more reliable judgment.

\subsection{Reprofile: Local Recursive Self-Improvement}

The recoverable capability frontier is policy-dependent rather than a fixed property of the candidate set. After Solver-RLVR updates the policy from $\pi_t$ to $\pi_{t+1}$, the rollout distribution over the same candidate problems may change: previously frontier-adjacent problems may become reliably solved or less informative, while other problems may move closer to the updated capability boundary.

Candidate problems are therefore re-evaluated after each policy update rather than following a fixed ranking obtained before training. Let the candidate pool at stage $t$ be

\begin{equation}
\mathcal{Q}_t=\{q_1,\ldots,q_M\}.
\end{equation}

At stage $t$, candidate problems are re-sampled under the current policy $\pi_t$, and the rollout statistics required by the frontier-selection score are recomputed using verifier-confirmed outcomes. The next training target is selected using the same top-1 rule defined in Consolidate:
\begin{equation}
q_t^\star
=
\arg\max_{q\in\mathcal{Q}_t}
S(q;\pi_t),
\end{equation}
where $S(q;\pi_t)$ denotes the frontier-selection score evaluated under the current policy. Because the score depends on the policy-induced rollout distribution, the ranking of candidate problems is recomputed after each policy update.

Training on $q_t^\star$ produces the updated policy $\pi_{t+1}$, under which the candidate pool is reprofiled before the subsequent selection. Repeating this profile--select--update cycle yields the policy-adaptive sequential RLVR process in ReSI-Math, where each policy update changes the evidence used to determine the next training target.









% ...

% ...

% ...

% ...

\section{Experiment}

\begin{table*}[t]
\centering
\small
\caption{
Main results of Solver.
The concrete selected problems and their selection-time profiles are reported in Table~\ref{tab:selected_problem_mapping}, with the corresponding problems for the three backbones visualized in Figures~\ref{fig:selected_2b_problems}--\ref{fig:selected_8b_problems}.
}
\label{tab:solver_main_backbones}

\begin{adjustbox}{max width=\textwidth}
\begin{tabular}{lllccccccccc}

\specialrule{.16em}{0pt}{.65ex}

\multirow{2}{*}{\textbf{Backbone}}
& \multirow{2}{*}{\textbf{Method}}
& \multirow{2}{*}{\textbf{Selected Problems}}
& \multirow{2}{*}{\textbf{Ckpt}}
& \multirow{2}{*}{\textbf{MATH500}}
& \multicolumn{4}{c|}{\textbf{WE-MATH testmini}}
& \multirow{2}{*}{\textbf{AIME 2025}}
& \multirow{2}{*}{\textbf{OlympiadBench}}
& \multirow{2}{*}{\textbf{Avg.}} \\

\cmidrule(lr){6-9}

&
&
&
&
&
\textbf{Overall}
& \textbf{1-step}
& \textbf{2-step}
& \textbf{3-step}
&
&
& \\

\specialrule{.05em}{.4ex}{.65ex}

\multicolumn{12}{c}{\textit{\textbf{Qwen3-VL-2B-Instruct}}} \\
\midrule

Base
& --
& --
& --
& 72.60 / --
& 54.14 / --
& 60.58 / --
& 44.44 / --
& 27.88 / --
& \underline{16.67 / --}
& 17.06 / --
& 40.12 / -- \\

Ours-Q1
& policy-adaptive
& $q_1^{(2B)}$
& 250
& 72.40 / -0.20
& 54.48 / +0.34
& 61.48 / +0.90
& 41.11 / -3.33
& 32.12 / +4.24
& \textbf{16.67 / +0.00}
& 16.77 / -0.29
& 40.08 / -0.04 \\

Ours-Q2
& policy-adaptive
& $q_1^{(2B)} \rightarrow q_2^{(2B)}$
& 550
& 74.00 / +1.40
& 56.32 / +2.18
& 62.22 / +1.64
& 46.39 / +1.95
& 34.55 / +6.67
& 6.67 / -10.00
& 19.14 / +2.08
& 39.03 / -1.09 \\

Ours-Q3
& policy-adaptive
& $q_1^{(2B)} \rightarrow q_2^{(2B)} \rightarrow q_3^{(2B)}$
& 200
& \textbf{75.40 / +2.80}
& \textbf{58.10 / +3.96}
& \textbf{64.36 / +3.78}
& \textbf{46.67 / +2.23}
& \textbf{36.97 / +9.09}
& 13.33 / -3.34
& \textbf{22.26 / +5.20}
& \textbf{42.27 / +2.15} \\

Fixed-Seq
& fixed order
& fixed 69-problem pool
& best
& 74.40 / +1.80
& 54.83 / +0.69
& 61.07 / +0.49
& 45.56 / +1.12
& 29.09 / +1.21
& 10.00 / -6.67
& 20.33 / +3.27
& 39.89 / -0.23 \\

Full-Batch
& joint GRPO
& same 69-problem pool
& best
& 74.00 / +1.40
& 54.71 / +0.57
& 61.15 / +0.57
& 44.44 / +0.00
& 29.70 / +1.82
& 10.00 / -6.67
& 18.84 / +1.78
& 39.39 / -0.73 \\

\specialrule{.08em}{.6ex}{.35ex}
\multicolumn{12}{c}{\textit{\textbf{Qwen3-VL-4B-Instruct}}} \\
\midrule

Base
& --
& --
& --
& 85.40 / --
& 67.36 / --
& 71.93 / --
& 60.83 / --
& 47.88 / --
& 26.67 / --
& 36.05 / --
& 53.87 / -- \\

Ours-Q1
& policy-adaptive
& $q_1^{(4B)}$
& 300
& \textbf{85.80 / +0.40}
& 68.45 / +1.09
& \underline{73.58 / +1.65}
& 59.72 / -1.11
& 49.70 / +1.82
& 26.67 / +0.00
& 35.46 / -0.59
& 54.09 / +0.22 \\

Ours-Q2
& policy-adaptive
& $q_1^{(4B)} \rightarrow q_2^{(4B)}$
& 100
& 85.40 / +0.00
& 68.28 / +0.92
& 73.17 / +1.24
& \textbf{62.22 / +1.39}
& 45.45 / -2.43
& \textbf{30.00 / +3.33}
& 33.53 / -2.52
& 54.30 / +0.43 \\

Ours-Q3
& policy-adaptive
& $q_1^{(4B)} \rightarrow q_2^{(4B)} \rightarrow q_3^{(4B)}$
& 250
& 85.40 / +0.00
& \textbf{68.79 / +1.43}
& \textbf{73.58 / +1.65}
& 61.39 / +0.56
& 49.70 / +1.82
& 26.67 / +0.00
& 36.05 / +0.00
& 54.23 / +0.36 \\

Fixed-Seq
& fixed order
& fixed 32-problem pool
& 650
& 85.20 / -0.20
& 68.05 / +0.69
& 73.00 / +1.07
& 58.89 / -1.94
& \underline{51.52 / +3.64}
& 23.33 / -3.34
& 34.87 / -1.18
& 52.86 / -1.01 \\

Full-Batch
& joint GRPO
& same 32-problem pool
& best
& 84.60 / -0.80
& 68.39 / +1.03
& 73.50 / +1.57
& 58.89 / -1.94
& \textbf{51.52 / +3.64}
& \underline{30.00 / +3.33}
& \textbf{36.20 / +0.15}
& \textbf{54.80 / +0.93} \\

\specialrule{.08em}{.6ex}{.35ex}
\multicolumn{12}{c}{\textit{\textbf{Qwen3-VL-8B-Instruct}}} \\
\midrule

Base
& --
& --
& --
& 85.80 / --
& 71.61 / --
& 75.72 / --
& 64.72 / --
& 56.36 / --
& 23.33 / --
& 32.20 / --
& 53.23 / -- \\

Ours-Q1
& policy-adaptive
& $q_1^{(8B)}$
& 500
& \underline{86.80 / +1.00}
& 71.95 / +0.34
& 75.88 / +0.16
& 65.56 / +0.84
& \textbf{56.97 / +0.61}
& 23.33 / +0.00
& \textbf{34.57 / +2.37}
& 54.16 / +0.93 \\

Ours-Q2
& policy-adaptive
& $q_1^{(8B)} \rightarrow q_2^{(8B)}$
& 500
& 86.00 / +0.20
& 73.05 / +1.44
& \textbf{77.70 / +1.98}
& 65.00 / +0.28
& 56.36 / +0.00
& 23.33 / +0.00
& 31.60 / -0.60
& 53.50 / +0.27 \\

Ours-Q3
& policy-adaptive
& $q_1^{(8B)} \rightarrow q_2^{(8B)} \rightarrow q_3^{(8B)}$
& 650
& \textbf{86.80 / +1.00}
& \textbf{73.51 / +1.90}
& 77.45 / +1.73
& \textbf{68.06 / +3.34}
& 56.36 / +0.00
& \textbf{26.67 / +3.34}
& 33.09 / +0.89
& \textbf{55.02 / +1.79} \\

Fixed-Seq
& fixed order
& fixed 28-problem pool
& best
& \underline{86.80 / +1.00}
& 72.13 / +0.52
& 76.63 / +0.91
& 64.44 / -0.28
& 55.76 / -0.60
& 16.67 / -6.66
& 32.79 / +0.59
& 52.10 / -1.13 \\

Full-Batch
& joint GRPO
& same 28-problem pool
& 20
& 85.40 / -0.40
& 71.44 / -0.17
& 75.88 / +0.16
& 64.44 / -0.28
& 53.94 / -2.42
& \underline{26.67 / +3.34}
& 32.79 / +0.59
& 54.08 / +0.85 \\

\specialrule{.16em}{0pt}{.65ex}

\end{tabular}
\end{adjustbox}
\end{table*}

In this section, we evaluate the effectiveness and generalizability of our ReSI-Math framework across multiple standard benchmarks.

\label{sec:experiment_exp_setup}

\subsection{Experimental Setup}

\paragraph{Models.}
We use Qwen3-VL-8B-Instruct as the primary backbone throughout our experiments, as well as Qwen3-VL-2B-Instruct and Qwen3-VL-4B-Instruct.
Qwen3-VL is an open-weight vision-language model family with strong text understanding, visual perception, and multimodal reasoning capabilities.

\paragraph{Datasets.}
We construct our evolutionary training pool from MathBook-Pro in WE-MATH 2.0 \citep{qiao2026wemath2}.
We construct the training data for both Solver-RLVR and Judge-RLVR from a common source of 243 seed mathematical problems. For Solver-RLVR, the evolutionary search produces a 28-problem candidate pool, from which one problem is selected at each sequential stage for single-problem RLVR.
For Judge-RLVR, trajectory-level supervision is derived from the same evolutionary source using verifier-labeled solver trajectories.
Detailed data construction, filtering criteria, and statistics are provided in Appendix~\ref{app:data_details}. For Solver checkpoint selection, we construct DEV192 by sampling 96 problems from MATH500 and 96 problems from WE-MATH testmini. Since DEV192 is contained in the full evaluation benchmarks reported in the main tables, we additionally evaluate on subsets that exclude these development examples. The corresponding results are reported in Appendix~\ref{app:dev_excluded}.


\paragraph{Benchmarks.} For Solver evaluation, we use four mathematical reasoning benchmarks: WE-MATH testmini\cite{qiao2025wemath-testmini}, MATH500\cite{hendrycks2021math500}, AIME 2025\cite{aime25}, and OlympiadBench\cite{he2024olympiadbench}. 
For Judge evaluation, we use JudgeBench \citep{tan2025judgebench}, a pairwise LLM-as-a-Judge benchmark with objectively verifiable preference labels across Knowledge, Reasoning, Math, and Coding.
We use the GPT-4o split containing 350 response pairs.

\paragraph{Implementation Details.}
Both Solver-RLVR and Judge-RLVR use Group Relative Policy Optimization (GRPO)~\citep{shao2024deepseekmath} with LoRA adaptation.
Solver-RLVR uses 16 online rollouts per training problem with batch size 1, while Judge-RLVR uses 8 rollouts per candidate prompt. Full optimization hyperparameters are provided in Appendix~\ref{app:rlvr_hparams}.

\subsection{Main Results}
\label{sec:main_results}

\begin{table*}[t]
\centering
\small
\caption{Cross-scale summary of post-hoc rollout dynamics during policy-adaptive sequential Solver-RLVR. 
Full checkpoint-wise profiles are provided in Appendix~\ref{app:rollout_dynamics}.
}
\label{tab:rollout_dynamics_summary}
\begin{adjustbox}{max width=\textwidth}
\begin{tabular}{lccccccc}
\toprule
\textbf{Backbone}
& \textbf{Frontier Pool}
& \textbf{Q1 Target}
& \textbf{Q2 Target}
& \textbf{Q3 Target}
& \textbf{Q1 / Q2 at Q3-best}
& \textbf{Avg. $\Delta$}
& \textbf{Dynamics} \\
\midrule

Qwen3-VL-2B
& 69
& \rC{11/16} $\rightarrow$ \rA{\textbf{16/16}}
& \rE{2/16} $\rightarrow$ \rC{\textbf{8/16}}
& \rD{6/16} $\rightarrow$ \rB{\textbf{14/16}}
& \rA{16/16}\ /\ \rC{9/16}
& +2.15
& Stable retention \\

Qwen3-VL-4B
& 32
& \rD{5/16} $\rightarrow$ \rE{\textbf{1/16}}
& \rC{10/16} $\rightarrow$ \rD{\textbf{6/16}}$^{\dagger}$
& \rC{9/16} $\rightarrow$ \rB{\textbf{15/16}}
& \rE{0/16}\ /\ \rC{9/16}
& +0.36
& Target degradation \\

Qwen3-VL-8B
& 28
& \rD{6/16} $\rightarrow$ \rA{\textbf{16/16}}
& \rD{5/16} $\rightarrow$ \rA{\textbf{16/16}}
& \rD{7/16} $\rightarrow$ \rA{\textbf{16/16}}
& \rC{11/16}\ /\ \rA{16/16}
& +1.79
& Forgetting / recovery \\

\bottomrule
\end{tabular}
\end{adjustbox}

\end{table*}

\begin{table*}[t]
\centering
\caption{
Ablation on training-problem selection for single-problem RLVR with
Qwen3-VL-8B-Instruct.
}
\label{tab:ablation_random_selection}
\begin{adjustbox}{max width=\textwidth}
\begin{tabular}{llccccccccc}
\specialrule{.16em}{0pt}{.65ex}

\textbf{Method}
& \textbf{Selection}
& \textbf{Ckpt}
& \textbf{MATH500}
& \multicolumn{4}{c}{\textbf{WE-MATH testmini}}
& \textbf{AIME 2025}
& \textbf{OlympiadBench}
& \textbf{Avg.} \\

\cmidrule(lr){5-8}

&
&
&
&
\textbf{Overall}
& \textbf{1-step}
& \textbf{2-step}
& \textbf{3-step}
&
&
& \\

\specialrule{.05em}{.4ex}{.65ex}

Base
& --
& --
& 85.80
& 71.61
& 75.72
& 64.72
& 56.36
& 23.33
& 32.20
& 53.23 \\

Random-1
& random
& 400
& 85.20
& 71.78
& 76.87
& 64.44
& 50.30
& 23.33
& 32.94
& 53.31 \\

\textbf{Ours-Q1}
& \textbf{policy-adaptive}
& \textbf{500}
& \textbf{86.80}
& \textbf{71.95}
& \textbf{75.88}
& \textbf{65.56}
& \textbf{56.97}
& \textbf{23.33}
& \textbf{34.57}
& \textbf{54.16} \\

\specialrule{.16em}{0pt}{.65ex}
\end{tabular}
\end{adjustbox}
\end{table*}

\paragraph{Recoverability under Fixed-Policy Search.} Under the evolutionary search protocol, 621 correct--failed ($C$--$F$) crossover offspring are generated from the 243-problem source set, of which 272 are verified as correct, corresponding to an offspring-level recovery rate of 43.8\%. This result provides direct empirical evidence that failed trajectories can contain complementary reasoning information that remains recoverable under a fixed policy.

\paragraph{Solver Results across Model Scales.} Table~\ref{tab:solver_main_backbones} reports the policy-adaptive Solver-RLVR results across Qwen3-VL-2B, 4B, and 8B. The largest relative improvement is obtained on the 2B backbone: after three sequentially selected problems, MATH500 increases from 72.60 to 75.40 ($+2.80$), WE-MATH testmini from 54.14 to 58.10 ($+3.96$), and the four-benchmark average from 40.12 to 42.27 ($+2.15$). The corresponding average gains are $+0.36$ for 4B and $+1.79$ for 8B. Although the strongest relative gain is observed on 2B, the controlled scheduling ablations and mechanism analyses are conducted on the 8B backbone to maintain a consistent primary experimental setting.

For the matched scheduling baselines, \textbf{Fixed-Seq} removes the frontier-guided top-1 selection mechanism and traverses the entire candidate pool in a fixed order, training on one problem at a time with a training batch size of 1. \textbf{Full-Batch} instead jointly optimizes all problems in the corresponding candidate pool within each training batch. Full-Batch GRPO is evaluated under the same rollout budget as the policy-adaptive setting, while Fixed-Sequential is included as a lower-budget fixed-order reference; detailed budget accounting is provided in Appendix~\ref{app:rollout_budget}.

Under this controlled comparison, policy-adaptive sequential RLVR consistently outperforms Fixed-Seq across all three model scales. Relative to the corresponding Base models, policy-adaptive training improves the average score by 2.15 points on 2B, 0.36 points on 4B, and 1.79 points on 8B, whereas Fixed-Seq decreases the average by 0.23, 1.01, and 1.13 points, respectively.
These results indicate that dynamic selection is consistently more effective than traversing the same candidate pool in a fixed sequential order, while its advantage over joint optimization is model-dependent; the distinct 4B behavior is further examined through the rollout dynamics in Table~\ref{tab:rollout_dynamics_summary}.

\paragraph{Scale-Dependent and Non-Monotonic Frontier Dynamics.} Table~\ref{tab:rollout_dynamics_summary} reveals substantial scale dependence in both frontier size and training dynamics. The candidate pools contain 69, 32, and 28 problems for Qwen3-VL-2B, 4B, and 8B, respectively, suggesting that a larger fraction of the shared problem distribution lies near the recoverable capability boundary for smaller models. The rollout profiles further show that single-problem RLVR does not induce monotonic capability acquisition: 2B largely preserves previously acquired behavior, 4B exhibits target-specific degradation, and 8B shows temporary forgetting followed by partial recovery. Additional re-learning experiments further reveal reverse interference, where improving a target capability can suppress previously consolidated capabilities and, in more severe cases, induce partial forgetting. Detailed checkpoint-wise rollout profiles and reverse-interference analyses are provided in Appendix~\ref{app:rollout_dynamics}.

\subsubsection{Self-Judge Provides Structured but Imperfect Search Signals}
\label{sec:selfjudge_frontier}

Evolutionary search requires more than generating additional trajectories:
it also requires a mechanism for deciding which reasoning paths are
promising enough to preserve, compare, or recombine.
In our framework, the same model is switched from the Solver role to the
Self-Judge role to provide trajectory-level feedback during evolution.

However, Self-Judge is not treated as a correctness oracle.
Comparing blind judgments against verifier-confirmed outcomes exposes
two characteristic types of self-evaluation error.
We identify 33 false-positive cases in which an actually failed parent
trajectory is judged as correct, and 35 false-negative cases in which a
verified-correct child trajectory is judged as incorrect.
These errors provide 68 explicit self-correction signals for downstream
Judge training.

\begin{table}[t]
\centering
\caption{
Self-Judge errors identified from verifier-confirmed evolutionary
trajectories.
}
\label{tab:selfjudge_errors}
\resizebox{\columnwidth}{!}{
\begin{tabular}{lcccc}
\specialrule{.16em}{0pt}{.65ex}
\textbf{Type}
& \textbf{Trajectory}
& \textbf{Verifier}
& \textbf{Blind Judge}
& \textbf{Count} \\
\specialrule{.05em}{.4ex}{.65ex}

False Positive
& Parent
& F
& C
& 33 \\

False Negative
& Child
& C
& F
& 35 \\

\specialrule{.16em}{0pt}{.65ex}
\end{tabular}
}
\end{table}

These observations suggest that Self-Judge provides structured information
about trajectory quality, but remains imperfect.
We therefore use it as a \emph{search controller} rather than as the final
source of correctness.
The Self-Judge determines which trajectories deserve attention during
search, while the deterministic verifier independently confirms whether
the resulting outcome is correct.

A further consequence is that evolutionary search exposes not only
reasoning failures, but also the model's own evaluation failures.
This creates a natural source of trajectory-level supervision for
Judge-RLVR, which we examine below.



\paragraph{Solver Results.} Table~\ref{tab:solver_main_backbones} summarizes the policy-adaptive Solver-RLVR results across Qwen3-VL-2B, 4B, and 8B, together with matched scheduling baselines. The largest relative improvement is obtained on the 2B backbone, where the four-benchmark average increases from 40.12 to 42.27 ($+2.15$). The corresponding gains are $0.36$ points for 4B and $1.79$ for 8B. On the 8B backbone, which is used for the main controlled ablations, policy-adaptive sequential training outperforms fixed-order sequential training and achieves a larger gain than joint optimization over the 28-problem candidate pool.

\begin{figure}[t]
    \centering
    \includegraphics[width=0.45\textwidth]{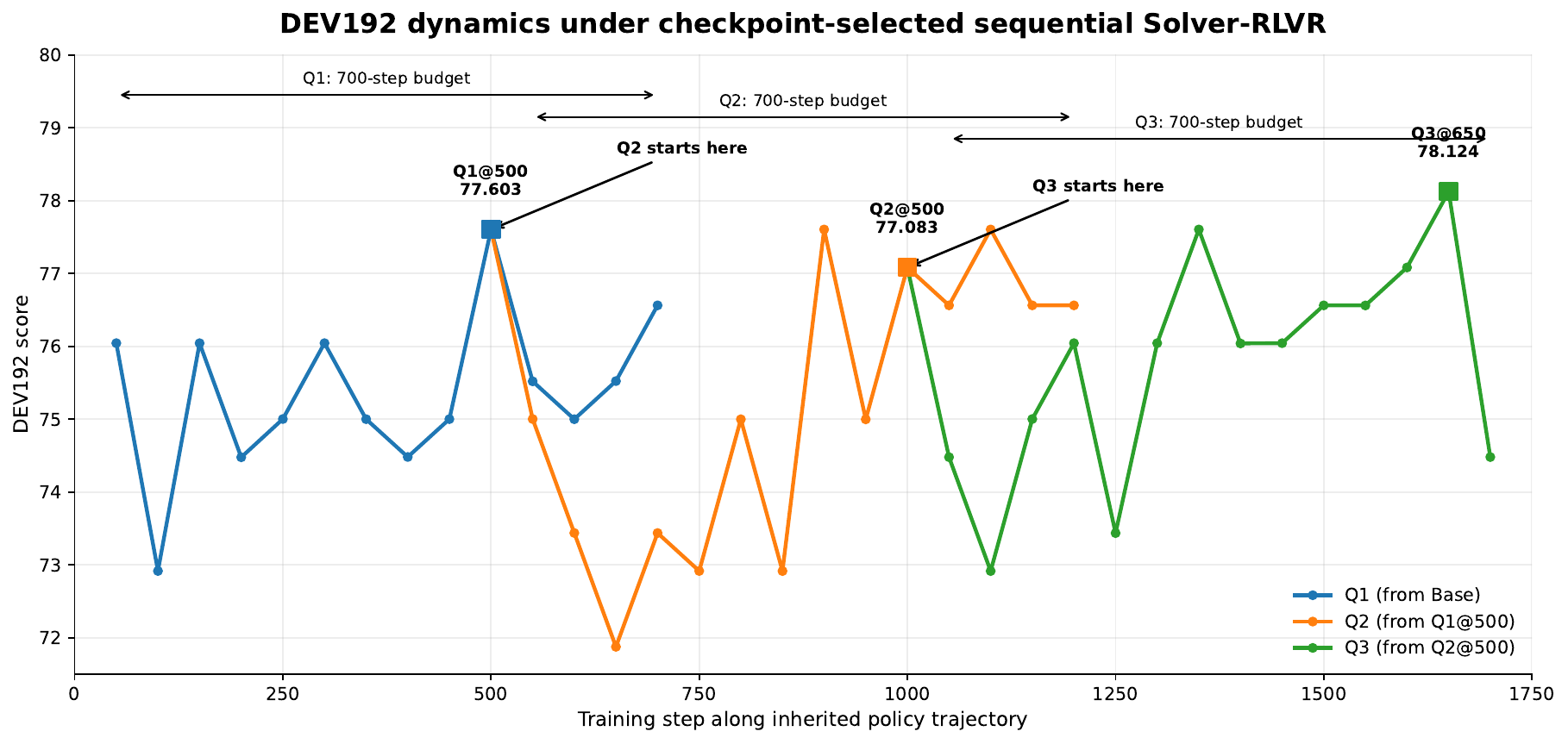}
    \caption{Policy-adaptive sequential Solver-RLVR dynamics on DEV192. 
    Q1, Q2, and Q3 are trained sequentially, with each stage initialized from the DEV192-selected checkpoint of the preceding stage, yielding Q1@500, Q2@500, and Q3@650. 
    }
    \label{fig:solver_dev_dynamics}
\end{figure}


\begin{table}[t]
\centering
\caption{
JudgeBench results of Judge-RLVR.
}
\label{tab:judge_main}
\resizebox{\columnwidth}{!}{
\begin{adjustbox}{max width=0.82\linewidth}
\begin{tabular}{lccccc}
\specialrule{.16em}{0pt}{.65ex}
\textbf{Model}
& \textbf{Knowledge}
& \textbf{Reasoning}
& \textbf{Math}
& \textbf{Coding}
& \textbf{Overall} \\
\specialrule{.05em}{.4ex}{.65ex}

Base
& 57.14
& \textbf{75.51}
& 80.36
& 61.90
& 66.57 \\

Judge-RLVR
& \textbf{59.09}
& 73.47
& \textbf{82.14}
& \textbf{64.29}
& \textbf{67.43} \\

\specialrule{.16em}{0pt}{.65ex}
\end{tabular}
\end{adjustbox}
}
\end{table}

\subsection{Ablation Study}





To examine whether the gains from single-problem RLVR can be attributed merely to training on an arbitrary example, we compare our policy-adaptive Q1 selection with a Random-1 baseline under the same 28-problem candidate pool, the same single-problem RLVR setting, and the same training budget. Random-1 improves the overall average only marginally, from 53.23 to 53.31 (+0.08), with inconsistent changes across evaluation dimensions: WE-MATH 1-step increases to 76.87, whereas MATH500 decreases from 85.80 to 85.20 and WE-MATH 3-step drops from 56.36 to 50.30. In contrast, frontier-guided Q1 raises the overall average to 54.16 (+0.93), outperforming Random-1 by 0.85 points, while also exceeding the Base model on MATH500, WE-MATH Overall, 2-step, 3-step, and OlympiadBench. These results indicate that, in the extreme low-data single-problem RLVR regime, the observed gains cannot be explained solely by training on an arbitrary example; under a matched training budget, the alignment between the selected problem and the current policy state materially affects downstream generalization performance.


\section{Conclusion}



We introduce \textbf{ReSI-Math}, a framework for recursive self-improvement in mathematical reasoning. Self-Judge-guided evolutionary search probes the recoverable capability frontier and uses crossover to repair failed reasoning into verifier-confirmed correct trajectories. Recoverable problems and judgment–verification discrepancies provide complementary supervision for Solver-RLVR and Judge-RLVR, respectively. Policy-adaptive sequential RLVR re-profiles the problem pool after each policy update to guide subsequent training. Experiments show that evolutionary search uncovers capabilities not yet reliably expressed by the model, while dynamic re-profiling outperforms static schedules in the extreme low-data regime. Thus, ReSI-Math extends evolutionary search beyond data synthesis to capability probing and learning-signal discovery, translating search-time recoverability into more reliable reasoning.


\clearpage
\section*{Limitations}




This work has two main limitations. First, our dynamic capability frontier is an operational, empirically defined notion under the current experimental setting. While sufficient for our analysis, it does not provide a strict characterization of the model's underlying capability boundary relative to the training data. Second, our experiments focus primarily on the Qwen3-VL family and mathematical reasoning, and we optimize the Solver and Judge separately rather than jointly. Generalization to other model families and reasoning domains, as well as the potential benefits of joint Solver--Judge optimization, remain important directions for future work.

\FloatBarrier
\bibliography{references}

\clearpage

\appendix
\section*{Appendix}
\addcontentsline{toc}{section}{Appendix}


\section{Training Data Construction}
\label{app:data_details}

\subsection{Solver-RLVR Data Construction}

The Solver receives a text problem together with its associated image when available; text-only benchmarks contain no image input.
The model generates a reasoning trajectory followed by a final answer formatted as
\verb|\boxed{...}|.
A deterministic rule-based verifier parses the final answer and compares it with the official reference answer.
The verifier supports multiple answer formats, including multiple-choice options, integers, decimals, fractions, radicals, coordinates, and symbolic expressions.
A rollout receives reward 1 if the final answer is verified as correct and 0 otherwise.

\subsection{Judge-RLVR Data Construction}

The Judge receives the original problem, the associated image when available, and a candidate Solver trajectory.
Instead of generating a new solution, it predicts whether the candidate reaches a verifier-confirmed correct outcome.
The final judgment is represented as
\verb|\boxed{1}| for correct and
\verb|\boxed{0}| for incorrect.
The underlying verifier label is hidden from the model during RLVR.
The Judge receives reward 1 if its prediction matches the hidden label and 0 otherwise.
Therefore, Judge-RLVR trains outcome discrimination rather than direct answer generation.

\subsection{Dataset Statistics}

We select 243 problems organized into such families as the seed pool for evolutionary search.
This family structure allows us to study complementary reasoning behaviors among closely related multimodal problems and to perform crossover between trajectories that share common semantic structure.

In our preprocessing, we group each base instance and its progressive variants into a common problem family, represented using the internal identifiers \texttt{base/x/y/z/xy/xz/yz/xyz}.

The resulting evolutionary trajectories provide two types of downstream supervision.
At the \emph{problem level}, a failure-to-success transition identifies a problem that lies near the model's recoverable capability frontier and can therefore serve as a candidate for Solver-RLVR.
At the \emph{trajectory level}, the evolutionary process produces verifier-confirmed correct and incorrect candidate solutions, which can be converted into Judge-RLVR examples.
Thus, the same evolutionary source provides complementary supervision for solving and judging, although the corresponding training units are different: problems for Solver-RLVR and candidate trajectories for Judge-RLVR.

\section{Implementation Details}
\label{app:implementation_details}

Table~\ref{tab:exp_setup} summarizes the functional differences among evolutionary search,
Solver-RLVR, and Judge-RLVR. Detailed configurations are provided in
the following subsections.

\begin{table*}[t]
\centering
\caption{
Functional comparison of evolutionary search, Solver-RLVR, and Judge-RLVR in ReSI-Math.
}
\label{tab:exp_setup}
\begin{adjustbox}{max width=0.90\linewidth}
\begin{tabular}{lccc}
\specialrule{.16em}{0pt}{.65ex}

\textbf{Setting}
& \textbf{Evolution}
& \textbf{Solver-RLVR}
& \textbf{Judge-RLVR} \\
\specialrule{.05em}{.4ex}{.65ex}

Primary role
& Frontier discovery
& Solver optimization
& Judge optimization \\

Basic unit
& trajectory
& problem
& candidate trajectory \\

Sampling granularity
& 16 trajectories (Round 0)
& 16 rollouts
& 8 rollouts \\

Search / optimization
& crossover
& GRPO
& GRPO \\

Iterations
& 3 rounds
& 700 steps / stage
& -- \\

External signal
& Self-Judge analysis + verifier outcome
& verifier reward
& verifier-guided judgment label \\

Policy parameters updated
& No
& Solver policy
& Judge policy \\

Reference answer exposed
& No
& No
& No \\

\specialrule{.16em}{0pt}{.65ex}
\end{tabular}
\end{adjustbox}
\end{table*}

\subsection{Rule-Based Verifier}

We use a deterministic rule-based outcome verifier as the external correctness anchor.
During blind crossover and blind self-judgment, the model is not given the reference answer, verifier label, or information about which parent trajectory is correct.
After generation, the verifier checks the final outcome and assigns a correctness label.
The verifier-confirmed outcome can subsequently be used by the guided Judge branch and as the trusted binary supervision for RLVR, while the reference answer itself remains hidden from the Judge.

Conceptually, the Self-Judge determines \emph{where to search}, whereas the rule-based verifier confirms \emph{whether the resulting trajectory succeeds}.

\subsection{Evolutionary Search Configuration}
\label{app:evolution}

We use the v7-era Self-Judge-guided evolutionary framework throughout the main experiments.
For each seed problem, the Solver first generates 16 independent trajectories in Round~0.
A rule-based verifier then determines their outcomes, after which we maintain a fixed population of four trajectories consisting of two verifier-confirmed correct trajectories and two failed trajectories.

The population is evolved for three rounds.
At each round, candidate parent trajectories are selected based on the current population and Self-Judge feedback, and crossover is used to generate new candidate trajectories.
The offspring are verified and the population is restored to the fixed $2\mathrm{C}+2\mathrm{F}$ composition.
A Blind Judge and a Guided Judge then analyze trajectory correctness, failure modes, and complementary information to support the subsequent round.
The Blind Judge does not observe verifier outcomes, whereas the Guided Judge may use verifier-confirmed outcome feedback without observing the reference solution.

\subsection{RLVR Hyperparameters}
\label{app:rlvr_hparams}

Table~\ref{tab:rlvr_hparams} summarizes the optimization hyperparameters used for the two training objectives.

For Solver-RLVR, we set the rollout size to 16 to match the 16-way sampling used in evolutionary search and subsequent policy profiling. This keeps the sampling granularity consistent when estimating whether a problem is currently all-fail, mixed, or all-pass and when optimizing the Solver on the selected problem.

Table~\ref{tab:rlvr_hparams} summarizes the main optimization
hyperparameters used for Solver-RLVR and Judge-RLVR.
Both variants use GRPO with LoRA adaptation, while differing primarily
in rollout size, batch size, and maximum response length.

\begin{table*}[t]
\centering
\caption{Training hyperparameters for Solver-RLVR and Judge-RLVR.}
\label{tab:rlvr_hparams}

\small
\begin{tabular}{lcc@{\hspace{1.2cm}}lcc}
\specialrule{.16em}{0pt}{.65ex}

\textbf{Configuration}
& \textbf{Solver}
& \textbf{Judge}
&
\textbf{Configuration}
& \textbf{Solver}
& \textbf{Judge} \\

\specialrule{.05em}{.4ex}{.65ex}

Algorithm
& GRPO & GRPO
&
Max prompt length
& 6144 & 6144 \\

Train batch size
& 1 & 4
&
Max response length
& 2048 & 1024 \\

Rollout $n$
& \textbf{16} & 8
&
Max model length
& 8192 & 8192 \\

Temperature
& 0.7 & 0.7
&
PPO mini-batch size
& 1 & 4 \\

Top-$p$
& 1.0 & 1.0
&
PPO micro-batch / GPU
& 1 & 1 \\

Learning rate
& $3\times10^{-6}$ & $3\times10^{-6}$
&
KL in reward
& No & No \\

LoRA rank
& 8 & 8
&
KL loss
& No & No \\

LoRA alpha
& 16 & 16
&
Entropy coefficient
& 0 & 0 \\

&
&
&
Save frequency
& Every 50 steps & -- \\

\specialrule{.16em}{0pt}{.65ex}
\end{tabular}
\end{table*}

\section{Robustness to DEV192 Checkpoint-Selection Overlap}
\label{app:dev_excluded}

Our main results report performance on the complete MATH500 and
WE-MATH testmini benchmarks for direct comparability with prior work.
DEV192, which consists of 96 examples from each benchmark, is used
only for checkpoint selection and never contributes to parameter
updates or training-data construction.
Since these examples remain part of the full benchmark evaluation,
checkpoint selection may introduce a small optimistic selection bias.
We therefore additionally report performance after removing all
DEV192 examples from the corresponding benchmarks.

Specifically, removing the 96 development examples yields
MATH404 and WE-MATH1644.
The last column reports the average accuracy over these two
DEV-excluded evaluation sets.

\section{Crossover Prompt}
\label{app:crossover_prompt}

We reproduce the crossover prompt used in our v7-era implementation.
The prompt is instantiated from a fixed system--user template with several
task-dependent fields.
The two parent trajectories are presented anonymously, and their order is
deterministically randomized for each question, operator, and replicate.
Correctness labels are used only for external task construction and are not
exposed in the blind crossover prompt.
The main evolutionary data use the blind arm, while the oracle-hinted arm
is included only as an additional control condition.

\begin{figure*}[t]
\centering
\begin{minipage}{0.96\textwidth}

\textbf{System Prompt}

\vspace{0.3em}

\begin{promptbox}
You are the blind Crossover role. You receive two anonymous candidate
trajectories. Never assume either candidate is correct. Re-evaluate the
problem from the image and question, combine only information you can
independently justify, and output one complete solution.
\end{promptbox}

\vspace{0.8em}

\textbf{User Prompt Template}

\vspace{0.3em}

\begin{promptbox}
{IMAGE}
Question:
{QUESTION}

Parent A:
{PARENT_A}

Parent B:
{PARENT_B}

{OPTIONAL_CONTROLLED_HINT}
Crossover instruction: {OPERATOR_INSTRUCTION}
Independent offspring sample index: {REPLICATE_INDEX}. Recheck the problem
and produce an independent sample rather than copying another offspring.
Return reasoning inside <think>...</think> and the final answer inside
<answer>...</answer>.
{FINAL_ANSWER_FORMAT_CONSTRAINT}
\end{promptbox}

\end{minipage}

\caption{
Crossover prompt template used in the v7-era evolutionary search.
Fields in braces are instantiated dynamically according to the crossover
task; their exact values are summarized in
Table~\ref{tab:crossover_dynamic_fields}.
}
\label{fig:crossover_prompt}
\end{figure*}

\begin{table*}[t]
\centering
\caption{
Dynamic fields instantiated in the crossover prompt.
The controlled hint is empty in the main blind arm and is populated only
for the oracle-hinted control condition.
}
\label{tab:crossover_dynamic_fields}

\begin{adjustbox}{max width=0.96\textwidth}
\begin{tabular}{
    >{\raggedright\arraybackslash}p{0.22\textwidth}
    >{\raggedright\arraybackslash}p{0.20\textwidth}
    >{\raggedright\arraybackslash}p{0.50\textwidth}
}
\specialrule{.16em}{0pt}{.65ex}

\textbf{Field}
& \textbf{Condition}
& \textbf{Instantiated Text} \\

\specialrule{.05em}{.4ex}{.65ex}

\multirow{2}{*}{
\makecell[l]{
\texttt{OPERATOR\_}\\
\texttt{INSTRUCTION}
}}
& Preserve--Repair
& \texttt{Preserve independently valid steps, locate contradictions,
repair them, and derive a complete solution.}
\\[0.8ex]

&
Independent Recombination
& \texttt{Rebuild the solution using a meaningfully different organization;
use both parents only as fallible evidence and recheck every key step.}
\\[1.2ex]

\multirow{2}{*}{
\makecell[l]{
\texttt{OPTIONAL\_}\\
\texttt{CONTROLLED\_HINT}
}}
& Main blind arm
& \textit{Empty.}
\\[0.8ex]

&
Oracle-hinted control
& \texttt{Controlled hint: Exactly one of the two parent trajectories has
a correct final result and the other has an incorrect final result.
You are not told which is which.}
\\[1.2ex]

\multirow{2}{*}{
\makecell[l]{
\texttt{FINAL\_ANSWER\_}\\
\texttt{FORMAT\_CONSTRAINT}
}}
& Multiple-choice
& \texttt{Final-answer format constraint: output exactly one option label
A--H inside <answer>...</answer>, for example <answer>A</answer>.
Do not place words, LaTeX wrappers, or option content inside that tag.}
\\[0.8ex]

&
Non-multiple-choice
& \texttt{Final-answer format constraint: put only the final mathematical
answer inside <answer>...</answer>.}
\\

\specialrule{.16em}{.65ex}{0pt}
\end{tabular}
\end{adjustbox}
\end{table*}

\clearpage

\makeatletter
\global\@colroom=\@colht
\global\vsize=\@colht
\makeatother

\section{Frontier Selection and Policy Dynamics}
\label{app:frontier_dynamics}

\subsection{Selected Frontier Problems}
\label{app:selected_problems}

\begin{figure*}[!t]
\centering

\begin{tcolorbox}[
    enhanced,
    colback=white,
    colframe=black!55,
    boxrule=0.55pt,
    arc=0mm,
    left=3mm,
    right=3mm,
    top=3mm,
    bottom=3mm
]

\frontierrow
{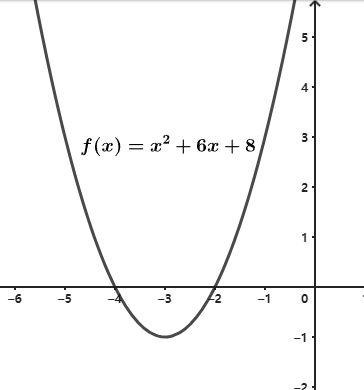}
{$q_1^{(2B)}$}
{family_302__x__row_2409}
{11/16}
{As shown in the figure, the value of the minimum of the function
multiplied by 2.5, plus 1.5, minus 2.5, and then multiplied by 50\% is.}

\vspace{0.7em}
\hrule
\vspace{0.7em}

\frontierrow
{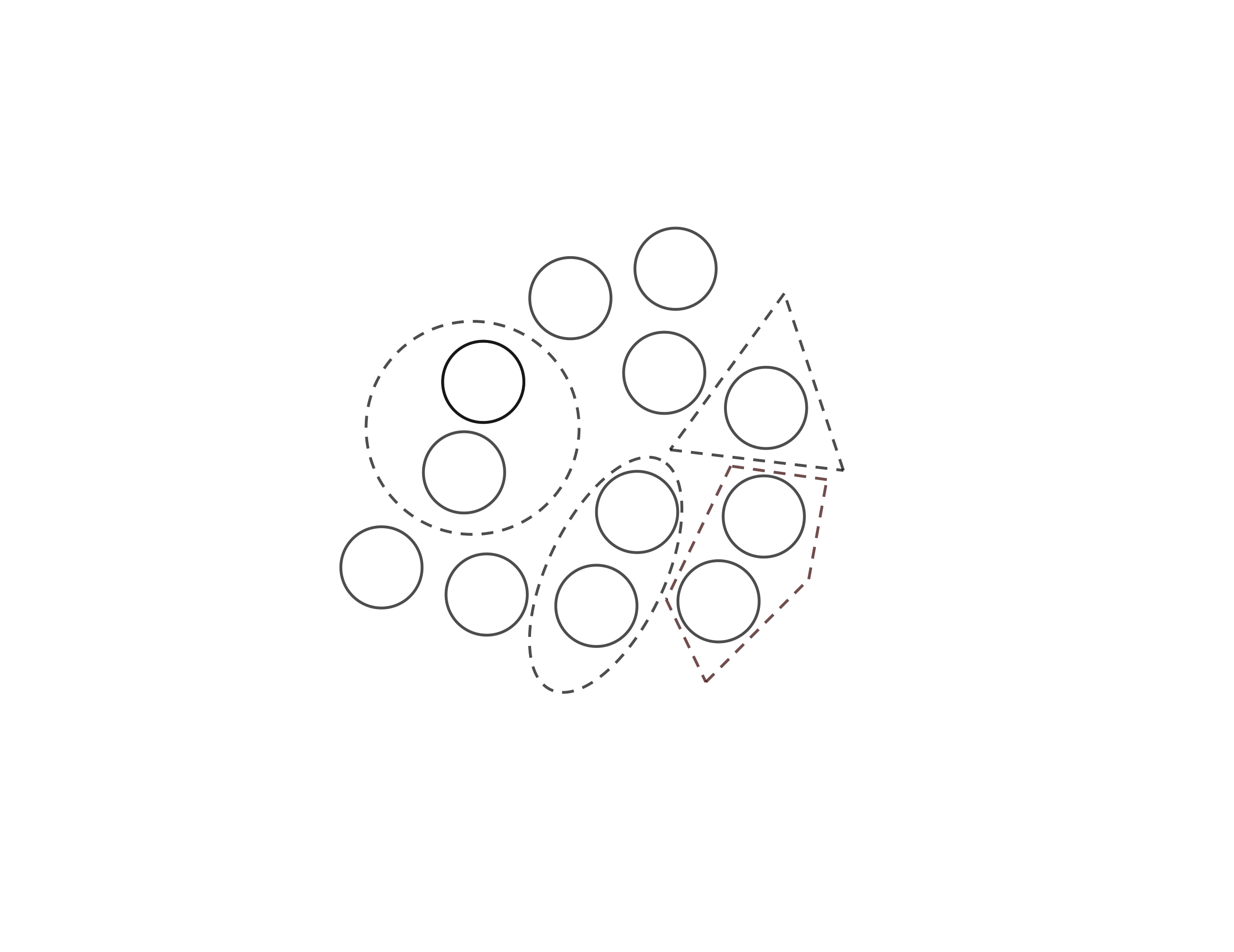}
{$q_2^{(2B)}$}
{family_320__y__row_2556}
{4/16}
{Given that there were originally 12 circles, how many circles remain
after removing the circles inside the dashed lines?
A.~3; B.~6; C.~5; D.~7.}

\vspace{0.7em}
\hrule
\vspace{0.7em}

\frontierrow
{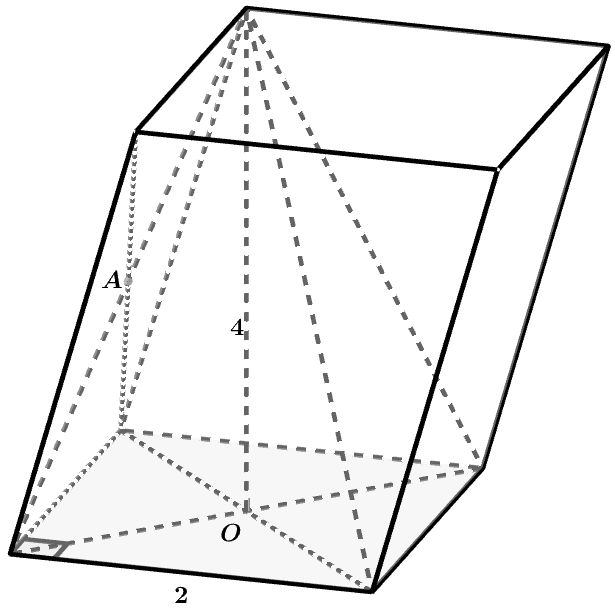}
{$q_3^{(2B)}$}
{family_420__yz__row_3357}
{7/16}
{As shown in the figure, the object is a prism.
The manufacturing cost of this object is 1 yuan per unit volume.
What is the total manufacturing cost of this object?
(Write only the number.)}

\end{tcolorbox}

\caption{
Concrete frontier problems selected for Qwen3-VL-2B.
Each row shows the visual input, original question, and
selection-time fixed-policy profile.
}
\label{fig:selected_2b_problems}
\end{figure*}

\begin{figure*}[!t]
\centering

\begin{tcolorbox}[
    enhanced,
    colback=white,
    colframe=black!55,
    boxrule=0.55pt,
    arc=0mm,
    left=3mm,
    right=3mm,
    top=3mm,
    bottom=3mm
]

\frontierrow
{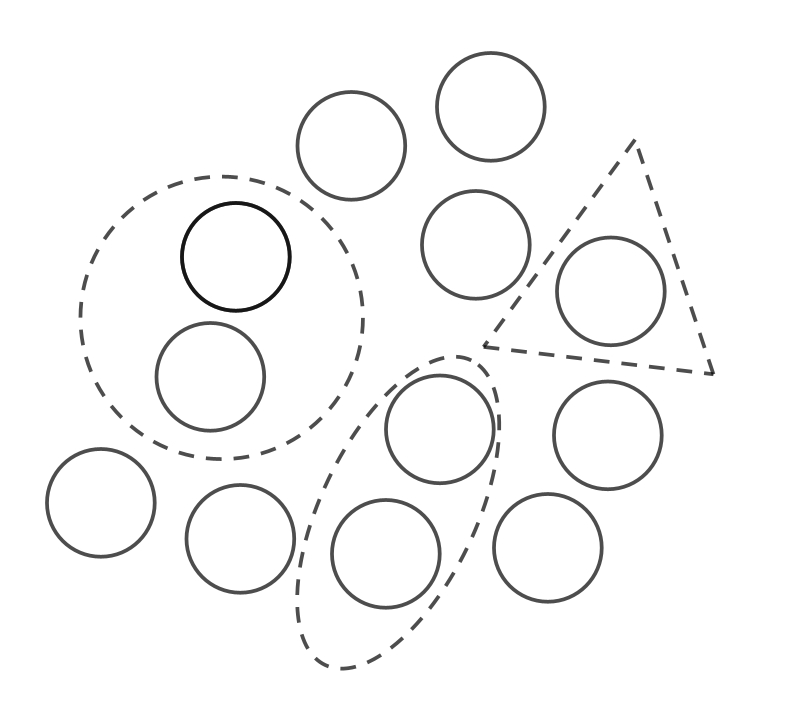}
{$q_1^{(4B)}$}
{family_320__xz__row_2559}
{5/16}
{There were originally 12 round cakes, as shown in the figure.
Now Mike has taken the cakes within the dashed lines.
Let $a$ be the number of cakes remaining, and let $b$ be the number
of extrema of the function $y=\frac{1}{3}x^3+x^2+2$.
What is the fourth term of a geometric sequence with first term $b$
and common ratio $a$?}

\vspace{0.7em}
\hrule
\vspace{0.7em}

\frontierrow
{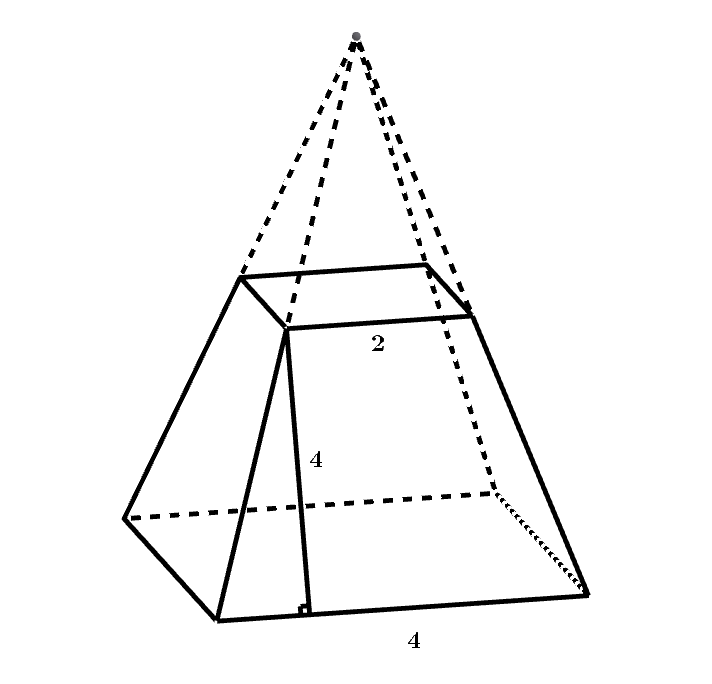}
{$q_2^{(4B)}$}
{family_436__y__row_3485}
{10/16}
{What is the area of the net of the following regular quadrilateral
frustum?}

\vspace{0.7em}
\hrule
\vspace{0.7em}

\frontierrow
{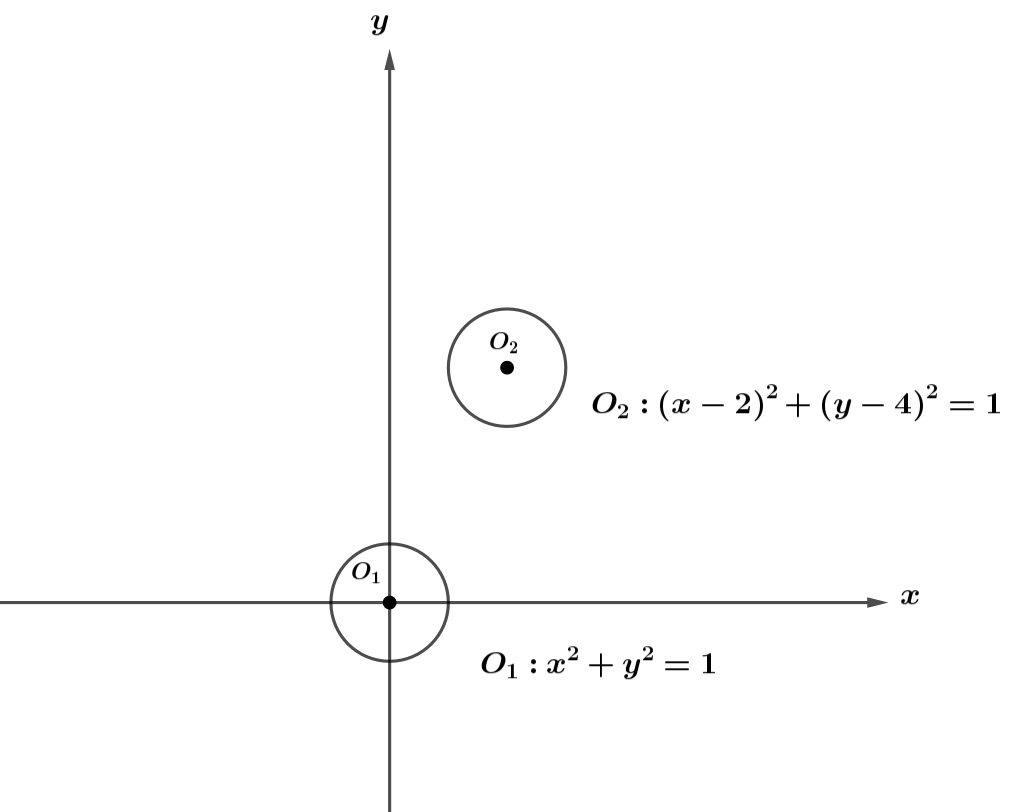}
{$q_3^{(4B)}$}
{family_203__z__row_1617}
{6/16}
{There are two disks of the same size on the table.
After fixing their positions, a straight wooden stick is used to touch
both disks and be tangent to both sides.
How many different ways are there to place the stick?
(Write only the number.)}

\end{tcolorbox}

\caption{
Concrete frontier problems selected for Qwen3-VL-4B.
Each row shows the visual input, original question, and
selection-time fixed-policy profile.
}
\label{fig:selected_4b_problems}
\end{figure*}

\begin{figure*}[!t]
\centering

\begin{tcolorbox}[
    enhanced,
    colback=white,
    colframe=black!55,
    boxrule=0.55pt,
    arc=0mm,
    left=3mm,
    right=3mm,
    top=3mm,
    bottom=3mm
]

\frontierrow
{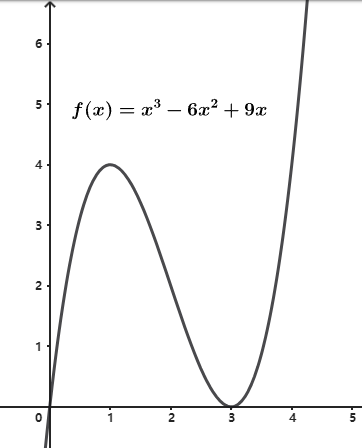}
{$q_1^{(8B)}$}
{family_311__xz__row_2485}
{6/16}
{A toy manufacturing company produces a new type of building block toy.
According to market research, the relationship between the daily
production quantity $x$ (in thousands of sets) and the profit per set
$y$ (in yuan) can be represented by the function shown in the figure.
The company's production planning department wants to understand the
fluctuation in profit.
What is the result of subtracting the minimum value of the function
from its maximum value?}

\vspace{0.7em}
\hrule
\vspace{0.7em}

\frontierrow
{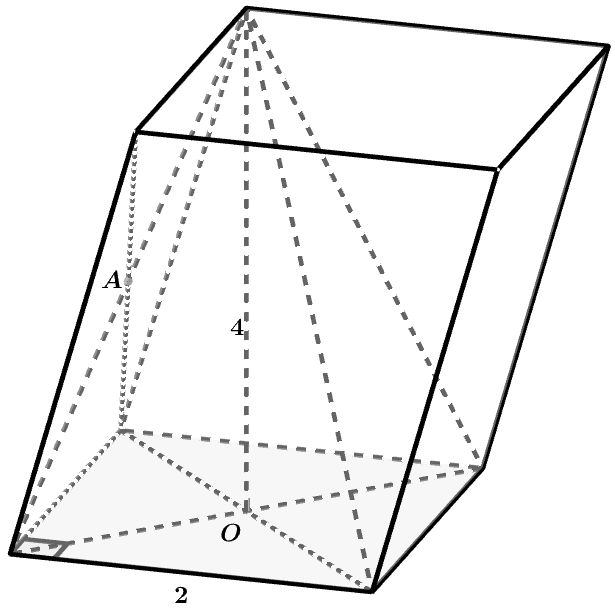}
{$q_2^{(8B)}$}
{family_420__y__row_3354}
{5/16}
{What is the volume of the prism shown in the figure?
(Enter only the number.)}

\vspace{0.7em}
\hrule
\vspace{0.7em}

\frontierrow
{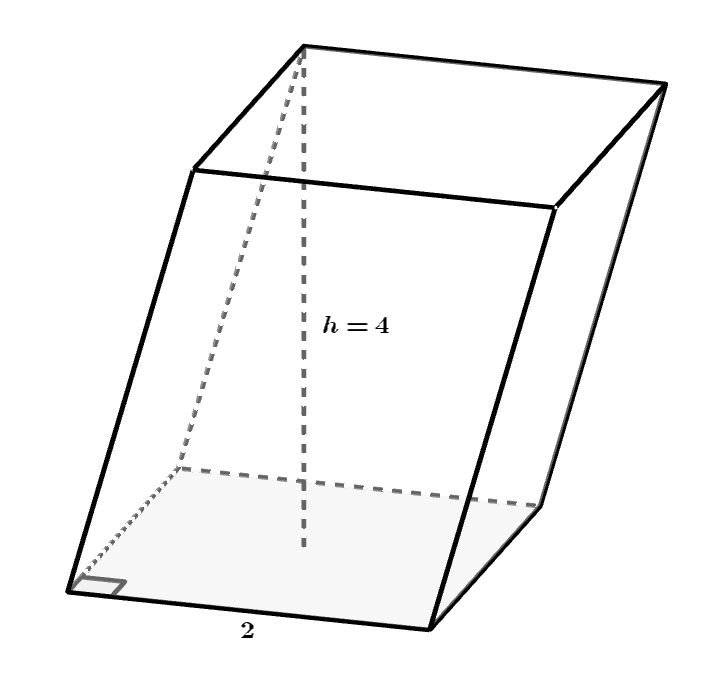}
{$q_3^{(8B)}$}
{family_420__base__row_3352}
{7/16}
{What is the volume of the prism shown in the figure?}

\end{tcolorbox}

\caption{
Concrete frontier problems selected for Qwen3-VL-8B.
Each row shows the visual input, original question, and
selection-time fixed-policy profile.
}
\label{fig:selected_8b_problems}
\end{figure*}

 \paragraph{Concrete Problem Instances.}
Figures~\ref{fig:selected_2b_problems}--\ref{fig:selected_8b_problems}
show the concrete Q1--Q3 problems selected for each backbone, including
the original question text and the corresponding visual input.

\subsection{Checkpoint-wise Rollout Dynamics of Sequential Solver-RLVR}
\label{app:rollout_dynamics}

%
%
\definecolor{accA}{RGB}{198,239,206}
\definecolor{accB}{RGB}{226,239,218}
\definecolor{accC}{RGB}{255,242,204}
\definecolor{accD}{RGB}{248,203,173}
\definecolor{accE}{RGB}{244,204,204}
\definecolor{selectedStep}{RGB}{221,235,247}
\definecolor{collapseStep}{RGB}{252,228,214}

 \paragraph{Frontier Size Varies with Model Scale.} 
The same family model induces substantially different frontier candidate pools across model scales. As shown in Table~\ref{tab:solver_main_backbones}, the resulting pools contain 69 problems for Qwen3-VL-2B, 32 for 4B, and 28 for 8B. This monotonic decrease with model size suggests that a larger fraction of the shared problem distribution lies near the recoverable capability boundary for smaller models, whereas stronger models move more problems away from this intermediate regime. The corresponding checkpoint-wise profiles in Tables~\ref{tab:2b_training_dynamics}--\ref{tab:8b_training_dynamics} further show that the behavior of these frontier problems is strongly scale-dependent.

 \paragraph{Frontier Training Exhibits Directional 
Plasticity.} 
Checkpoint-wise rollout profiles in Tables~\ref{tab:2b_training_dynamics}--\ref{tab:8b_training_dynamics}  reveal that single-problem RLVR does not induce uniformly monotonic capability improvement. On Qwen3-VL-2B (Table~\ref{tab:2b_training_dynamics}), Q1 reaches 16/16 early and remains fully retained throughout subsequent Q2 and Q3 stages, while Q3 is progressively consolidated to near-perfect rollout success; this stable retention is consistent with the largest relative benchmark gain observed in Table~\ref{tab:solver_main_backbones}. In contrast, Qwen3-VL-4B (Table~\ref{tab:4b_training_dynamics}) exhibits a target-specific degradation regime during Q2 training, where the post-hoc Q2 profile decreases from 10/16 at initialization to 0/16 at later checkpoints despite continued optimization on Q2. Qwen3-VL-8B (Table~\ref{tab:8b_training_dynamics}) shows an intermediate pattern, with temporary degradation of Q1 during Q2 training followed by partial recovery during Q3 training. These results indicate that frontier training is directionally plastic: a single problem can move the policy either toward or away from reliable expression of a targeted capability.

 \paragraph{Re-learning Can Induce Reverse Interference.} 
Table~\ref{tab:q4_reverse_interference} further shows that capability recovery is not necessarily globally beneficial. Starting from the selected Q3@650 policy on Qwen3-VL-8B, re-training Q1 increases its rollout success from 11/16 to 16/16, but simultaneously reduces the previously mastered Q2/Q3-related family from 64/64 to 23/64 and decreases DEV192 from 78.12 to 75.00. The corresponding full-test average also drops from 80.16 to 78.47. Thus, local reacquisition of a target capability can suppress previously consolidated capabilities, providing direct evidence that sequential single-problem RLVR exhibits non-monotonic and bidirectional interference. In more severe cases, this interference manifests as partial or temporary forgetting of previously acquired capabilities.

\subsection{Rollout-Budget Control}
\label{app:rollout_budget}
To separate the effect of training schedule from the amount of sampled supervision, we strictly match the rollout budget between Policy-Adaptive Sequential RLVR and Full-Batch GRPO. For all three model scales, sequential RLVR performs three stages of at most 700 single-problem updates with $n=16$ rollouts per update, yielding $3\times700\times16=33{,}600$ rollout responses. Full-Batch GRPO is assigned the same 2,100 prompt slots and therefore the same 33,600 rollout responses. Because the candidate-pool sizes differ across model scales, we use 30 full updates over the 69-problem pool followed by a 30-example partial update for 2B, 65 full updates over the 32-problem pool followed by a 20-example partial update for 4B, and 75 full updates over the 28-problem pool for 8B. The resulting budgets are exactly 33,600 rollouts in all cases. The Fixed-Sequential baseline follows the original 700-step fixed-order protocol with batch size 1 and $n=16$, corresponding to 11,200 rollouts; we therefore report it as a lower-budget scheduling reference rather than a budget-matched comparator.



\clearpage

\begin{table*}
\centering
\caption{
Concrete frontier problems selected during policy-adaptive sequential
Solver-RLVR.
``Selection Policy'' denotes the policy used to reprofile the candidate
pool before selecting the corresponding problem.
``Profile'' reports the number of verifier-confirmed correct trajectories
among 16 fixed-policy samples at selection time.
}
\label{tab:selected_problem_mapping}

\begin{adjustbox}{max width=0.96\textwidth}
\begin{tabular}{lccclc}
\specialrule{.16em}{0pt}{.65ex}

\textbf{Backbone}
& \textbf{Stage}
& \textbf{Symbol}
& \textbf{Selection Policy}
& \textbf{Selected Question ID}
& \textbf{Profile} \\

\specialrule{.05em}{.4ex}{.65ex}

\multirow{3}{*}{Qwen3-VL-2B}
& Q1
& $q_1^{(2B)}$
& Base
& \texttt{family\_302\_\_x\_\_row\_2409}
& 11/16 \\

& Q2
& $q_2^{(2B)}$
& Q1@250
& \texttt{family\_320\_\_y\_\_row\_2556}
& 4/16 \\

& Q3
& $q_3^{(2B)}$
& Q2@550
& \texttt{family\_420\_\_yz\_\_row\_3357}
& 7/16 \\

\midrule

\multirow{3}{*}{Qwen3-VL-4B}
& Q1
& $q_1^{(4B)}$
& Base
& \texttt{family\_320\_\_xz\_\_row\_2559}
& 5/16 \\

& Q2
& $q_2^{(4B)}$
& Q1@300
& \texttt{family\_436\_\_y\_\_row\_3485}
& 10/16 \\

& Q3
& $q_3^{(4B)}$
& Q2@100
& \texttt{family\_203\_\_z\_\_row\_1617}
& 6/16 \\

\midrule

\multirow{3}{*}{Qwen3-VL-8B}
& Q1
& $q_1^{(8B)}$
& Base
& \texttt{family\_311\_\_xz\_\_row\_2485}
& 6/16 \\

& Q2
& $q_2^{(8B)}$
& Q1@500
& \texttt{family\_420\_\_y\_\_row\_3354}
& 5/16 \\

& Q3
& $q_3^{(8B)}$
& Q2@500
& \texttt{family\_420\_\_base\_\_row\_3352}
& 7/16 \\

\specialrule{.16em}{.65ex}{0pt}
\end{tabular}
\end{adjustbox}
\end{table*}



\clearpage

\begin{table*}[!p]
\centering
\small
\setlength{\tabcolsep}{4.2pt}
\renewcommand{\arraystretch}{1.08}

\caption{
Checkpoint-wise rollout profiles during sequential single-problem Solver-RLVR on Qwen3-VL-2B.
Each entry reports the number of verifier-confirmed correct trajectories among 16 rollouts.
All tables use the same heatmap bins: 16/16, 13--15/16, 8--12/16, 4--7/16, and 0--3/16.
Blue step cells denote DEV192-selected checkpoints.Cell colors encode absolute rollout accuracy:dark green denotes 16/16, light green 13--15/16, yellow 8--12/16, orange 4--7/16, and light red 0--3/16. Blue step cells mark DEV192-selected checkpoints, while light-red step cells in the 4B Q2 stage indicate the observed optimization-collapse regime.
}
\label{tab:2b_training_dynamics}

\begin{adjustbox}{max width=\textwidth}
\begin{tabular}{cccc|cccc|cccc}
\toprule
\multicolumn{4}{c|}{\textbf{Q1 Training}} &
\multicolumn{4}{c|}{\textbf{Q2 Training}} &
\multicolumn{4}{c}{\textbf{Q3 Training}} \\
\cmidrule(lr){1-4}
\cmidrule(lr){5-8}
\cmidrule(lr){9-12}
\textbf{Step} & \textbf{Q1} & \textbf{Q2} & \textbf{Q3} &
\textbf{Step} & \textbf{Q1} & \textbf{Q2} & \textbf{Q3} &
\textbf{Step} & \textbf{Q1} & \textbf{Q2} & \textbf{Q3} \\
\midrule
0 & \cellcolor{accC}11/16 & \cellcolor{accE}1/16 & \cellcolor{accE}2/16 & 0 & \cellcolor{accA}16/16 & \cellcolor{accE}2/16 & \cellcolor{accD}6/16 & 0 & \cellcolor{accA}16/16 & \cellcolor{accC}8/16 & \cellcolor{accD}6/16 \\
50 & \cellcolor{accA}16/16 & \cellcolor{accE}3/16 & \cellcolor{accD}4/16 & 50 & \cellcolor{accA}16/16 & \cellcolor{accE}1/16 & \cellcolor{accE}2/16 & 50 & \cellcolor{accA}16/16 & \cellcolor{accC}10/16 & \cellcolor{accD}7/16 \\
100 & \cellcolor{accA}16/16 & \cellcolor{accE}3/16 & \cellcolor{accD}5/16 & 100 & \cellcolor{accA}16/16 & \cellcolor{accD}4/16 & \cellcolor{accD}5/16 & 100 & \cellcolor{accA}16/16 & \cellcolor{accC}11/16 & \cellcolor{accC}11/16 \\
150 & \cellcolor{accA}16/16 & \cellcolor{accD}5/16 & \cellcolor{accD}4/16 & 150 & \cellcolor{accA}16/16 & \cellcolor{accE}2/16 & \cellcolor{accE}2/16 & 150 & \cellcolor{accA}16/16 & \cellcolor{accC}12/16 & \cellcolor{accA}16/16 \\
200 & \cellcolor{accA}16/16 & \cellcolor{accE}2/16 & \cellcolor{accC}8/16 & 200 & \cellcolor{accA}16/16 & \cellcolor{accE}2/16 & \cellcolor{accD}6/16 & \cellcolor{selectedStep}\textbf{200} & \cellcolor{accA}\textbf{16/16} & \cellcolor{accC}\textbf{9/16} & \cellcolor{accB}\textbf{14/16} \\
\cellcolor{selectedStep}\textbf{250} & \cellcolor{accA}\textbf{16/16} & \cellcolor{accE}\textbf{2/16} & \cellcolor{accD}\textbf{6/16} & 250 & \cellcolor{accA}16/16 & \cellcolor{accE}3/16 & \cellcolor{accD}6/16 & 250 & \cellcolor{accA}16/16 & \cellcolor{accC}9/16 & \cellcolor{accB}15/16 \\
300 & \cellcolor{accA}16/16 & \cellcolor{accE}2/16 & \cellcolor{accD}7/16 & 300 & \cellcolor{accA}16/16 & \cellcolor{accE}3/16 & \cellcolor{accD}7/16 & 300 & \cellcolor{accA}16/16 & \cellcolor{accC}12/16 & \cellcolor{accA}16/16 \\
350 & \cellcolor{accA}16/16 & \cellcolor{accD}6/16 & \cellcolor{accD}7/16 & 350 & \cellcolor{accA}16/16 & \cellcolor{accE}3/16 & \cellcolor{accD}7/16 & 350 & \cellcolor{accA}16/16 & \cellcolor{accC}12/16 & \cellcolor{accA}16/16 \\
400 & \cellcolor{accA}16/16 & \cellcolor{accC}10/16 & \cellcolor{accD}5/16 & 400 & \cellcolor{accA}16/16 & \cellcolor{accD}5/16 & \cellcolor{accD}6/16 & 400 & \cellcolor{accA}16/16 & \cellcolor{accC}11/16 & \cellcolor{accA}16/16 \\
450 & \cellcolor{accA}16/16 & \cellcolor{accD}5/16 & \cellcolor{accD}6/16 & 450 & \cellcolor{accA}16/16 & \cellcolor{accD}5/16 & \cellcolor{accC}8/16 & 450 & \cellcolor{accA}16/16 & \cellcolor{accC}12/16 & \cellcolor{accA}16/16 \\
500 & \cellcolor{accA}16/16 & \cellcolor{accC}9/16 & \cellcolor{accD}4/16 & 500 & \cellcolor{accA}16/16 & \cellcolor{accC}8/16 & \cellcolor{accD}4/16 & 500 & \cellcolor{accA}16/16 & \cellcolor{accD}7/16 & \cellcolor{accA}16/16 \\
550 & \cellcolor{accA}16/16 & \cellcolor{accD}7/16 & \cellcolor{accD}4/16 & \cellcolor{selectedStep}\textbf{550} & \cellcolor{accA}\textbf{16/16} & \cellcolor{accC}\textbf{8/16} & \cellcolor{accD}\textbf{6/16} & 550 & \cellcolor{accA}16/16 & \cellcolor{accC}8/16 & \cellcolor{accA}16/16 \\
600 & \cellcolor{accA}16/16 & \cellcolor{accC}8/16 & \cellcolor{accD}4/16 & 600 & \cellcolor{accA}16/16 & \cellcolor{accC}11/16 & \cellcolor{accD}5/16 & 600 & \cellcolor{accA}16/16 & \cellcolor{accC}9/16 & \cellcolor{accA}16/16 \\
650 & \cellcolor{accA}16/16 & \cellcolor{accC}12/16 & \cellcolor{accD}4/16 & 650 & \cellcolor{accA}16/16 & \cellcolor{accC}12/16 & \cellcolor{accD}6/16 & 650 & \cellcolor{accA}16/16 & \cellcolor{accC}9/16 & \cellcolor{accA}16/16 \\
700 & \cellcolor{accA}16/16 & \cellcolor{accB}13/16 & \cellcolor{accE}3/16 & 700 & \cellcolor{accA}16/16 & \cellcolor{accB}13/16 & \cellcolor{accE}3/16 & 700 & \cellcolor{accA}16/16 & \cellcolor{accB}13/16 & \cellcolor{accA}16/16 \\
\bottomrule
\end{tabular}
\end{adjustbox}
\end{table*}

\begin{table*}[!p]
\centering
\small
\setlength{\tabcolsep}{4.2pt}
\renewcommand{\arraystretch}{1.08}

\caption{
Checkpoint-wise rollout profiles during sequential single-problem Solver-RLVR on Qwen3-VL-4B.
Each entry reports the number of verifier-confirmed correct trajectories among 16 rollouts.
All tables use the same heatmap bins: 16/16, 13--15/16, 8--12/16, 4--7/16, and 0--3/16.
Blue step cells denote DEV192-selected checkpoints.
Red step cells in Q2 training mark the observed optimization-collapse regime; accuracy cells retain the same heatmap encoding.
}
\label{tab:4b_training_dynamics}

\begin{adjustbox}{max width=\textwidth}
\begin{tabular}{cccc|cccc|cccc}
\toprule
\multicolumn{4}{c|}{\textbf{Q1 Training}} &
\multicolumn{4}{c|}{\textbf{Q2 Training}} &
\multicolumn{4}{c}{\textbf{Q3 Training}} \\
\cmidrule(lr){1-4}
\cmidrule(lr){5-8}
\cmidrule(lr){9-12}
\textbf{Step} & \textbf{Q1} & \textbf{Q2} & \textbf{Q3} &
\textbf{Step} & \textbf{Q1} & \textbf{Q2} & \textbf{Q3} &
\textbf{Step} & \textbf{Q1} & \textbf{Q2} & \textbf{Q3} \\
\midrule
0 & \cellcolor{accD}5/16 & \cellcolor{accC}11/16 & \cellcolor{accE}3/16 & 0 & \cellcolor{accE}2/16 & \cellcolor{accC}10/16 & \cellcolor{accD}5/16 & 0 & \cellcolor{accE}0/16 & \cellcolor{accC}8/16 & \cellcolor{accC}9/16 \\
50 & \cellcolor{accE}3/16 & \cellcolor{accC}8/16 & \cellcolor{accE}0/16 & 50 & \cellcolor{accE}2/16 & \cellcolor{accD}6/16 & \cellcolor{accD}4/16 & 50 & \cellcolor{accE}0/16 & \cellcolor{accD}6/16 & \cellcolor{accC}12/16 \\
100 & \cellcolor{accE}1/16 & \cellcolor{accC}9/16 & \cellcolor{accD}4/16 & \cellcolor{selectedStep}\textbf{100} & \cellcolor{accE}\textbf{0/16} & \cellcolor{accD}\textbf{6/16} & \cellcolor{accD}\textbf{5/16} & 100 & \cellcolor{accE}2/16 & \cellcolor{accD}7/16 & \cellcolor{accB}14/16 \\
150 & \cellcolor{accE}2/16 & \cellcolor{accC}8/16 & \cellcolor{accE}3/16 & 150 & \cellcolor{accE}1/16 & \cellcolor{accD}4/16 & \cellcolor{accB}13/16 & 150 & \cellcolor{accE}2/16 & \cellcolor{accC}8/16 & \cellcolor{accB}15/16 \\
200
& \cellcolor{accE}2/16
& \cellcolor{accC}8/16
& \cellcolor{accD}6/16
&
\collapsemark{200}
& \cellcolor{accE}0/16
& \collapsemark{0/16}
& \cellcolor{accB}13/16
&
200
& \cellcolor{accE}0/16
& \cellcolor{accC}9/16
& \cellcolor{accB}15/16
\\

250
& \cellcolor{accE}2/16
& \cellcolor{accC}9/16
& \cellcolor{accE}3/16
&
\collapsemark{250}
& \cellcolor{accE}0/16
& \collapsemark{0/16}
& \cellcolor{accB}14/16
&
250
& \cellcolor{accE}0/16
& \cellcolor{accC}9/16
& \cellcolor{accB}15/16
\\

300
& \cellcolor{accE}1/16
& \cellcolor{accD}5/16
& \cellcolor{accD}7/16
&
\collapsemark{300}
& \cellcolor{accE}0/16
& \collapsemark{0/16}
& \cellcolor{accC}12/16
&
300
& \cellcolor{accE}0/16
& \cellcolor{accC}8/16
& \cellcolor{accA}16/16
\\
350 & \cellcolor{accE}0/16 & \cellcolor{accD}7/16 & \cellcolor{accD}5/16 & 350 & -- & -- & -- & 350 & \cellcolor{accE}1/16 & \cellcolor{accC}10/16 & \cellcolor{accA}16/16 \\
400 & \cellcolor{accE}0/16 & \cellcolor{accE}0/16 & \cellcolor{accC}9/16 & 400 & -- & -- & -- & 400 & \cellcolor{accE}1/16 & \cellcolor{accC}8/16 & \cellcolor{accA}16/16 \\
450 & \cellcolor{accE}2/16 & \cellcolor{accE}0/16 & \cellcolor{accA}16/16 & 450 & -- & -- & -- & 450 & \cellcolor{accE}1/16 & \cellcolor{accC}11/16 & \cellcolor{accB}15/16 \\
500 & \cellcolor{accD}4/16 & \cellcolor{accE}1/16 & \cellcolor{accA}16/16 & 500 & -- & -- & -- & 500 & \cellcolor{accE}1/16 & \cellcolor{accC}8/16 & \cellcolor{accA}16/16 \\
550 & \cellcolor{accE}1/16 & \cellcolor{accE}0/16 & \cellcolor{accB}15/16 & 550 & -- & -- & -- & 550 & \cellcolor{accE}0/16 & \cellcolor{accC}8/16 & \cellcolor{accA}16/16 \\
600 & \cellcolor{accE}0/16 & \cellcolor{accE}0/16 & \cellcolor{accB}13/16 & 600 & -- & -- & -- & 600 & \cellcolor{accE}0/16 & \cellcolor{accD}7/16 & \cellcolor{accB}15/16 \\
650 & \cellcolor{accE}0/16 & \cellcolor{accE}0/16 & \cellcolor{accA}16/16 & 650 & -- & -- & -- & 650 & \cellcolor{accE}2/16 & \cellcolor{accC}8/16 & \cellcolor{accA}16/16 \\
700 & \cellcolor{accE}0/16 & \cellcolor{accE}0/16 & \cellcolor{accA}16/16 & 700 & -- & -- & -- & 700 & \cellcolor{accE}2/16 & \cellcolor{accC}11/16 & \cellcolor{accA}16/16 \\
\bottomrule
\end{tabular}
\end{adjustbox}
\end{table*}

\begin{table*}[!p]
\centering
\small
\setlength{\tabcolsep}{4.2pt}
\renewcommand{\arraystretch}{1.08}

\caption{
Checkpoint-wise rollout profiles during sequential single-problem Solver-RLVR on Qwen3-VL-8B.
Each entry reports the number of verifier-confirmed correct trajectories among 16 rollouts.
All tables use the same heatmap bins: 16/16, 13--15/16, 8--12/16, 4--7/16, and 0--3/16.
Blue step cells denote DEV192-selected checkpoints. 
}
\label{tab:8b_training_dynamics}

\begin{adjustbox}{max width=\textwidth}
\begin{tabular}{cccc|cccc|cccc}
\toprule
\multicolumn{4}{c|}{\textbf{Q1 Training}} &
\multicolumn{4}{c|}{\textbf{Q2 Training}} &
\multicolumn{4}{c}{\textbf{Q3 Training}} \\
\cmidrule(lr){1-4}
\cmidrule(lr){5-8}
\cmidrule(lr){9-12}
\textbf{Step} & \textbf{Q1} & \textbf{Q2} & \textbf{Q3} &
\textbf{Step} & \textbf{Q1} & \textbf{Q2} & \textbf{Q3} &
\textbf{Step} & \textbf{Q1} & \textbf{Q2} & \textbf{Q3} \\
\midrule
0 & \cellcolor{accD}6/16 & \cellcolor{accD}5/16 & \cellcolor{accD}4/16 & 0 & \cellcolor{accA}16/16 & \cellcolor{accD}5/16 & \cellcolor{accE}2/16 & 0 & \cellcolor{accC}8/16 & \cellcolor{accA}16/16 & \cellcolor{accD}7/16 \\
50 & -- & -- & -- & 50 & \cellcolor{accC}10/16 & \cellcolor{accD}6/16 & \cellcolor{accC}11/16 & 50 & \cellcolor{accC}11/16 & \cellcolor{accA}16/16 & \cellcolor{accB}15/16 \\
100 & -- & -- & -- & 100 & \cellcolor{accC}11/16 & \cellcolor{accC}9/16 & \cellcolor{accC}9/16 & 100 & \cellcolor{accC}9/16 & \cellcolor{accA}16/16 & \cellcolor{accB}15/16 \\
150 & -- & -- & -- & 150 & \cellcolor{accC}9/16 & \cellcolor{accB}13/16 & \cellcolor{accC}8/16 & 150 & \cellcolor{accC}9/16 & \cellcolor{accA}16/16 & \cellcolor{accA}16/16 \\
200 & -- & -- & -- & 200 & \cellcolor{accC}11/16 & \cellcolor{accB}15/16 & \cellcolor{accC}12/16 & 200 & \cellcolor{accC}11/16 & \cellcolor{accA}16/16 & \cellcolor{accA}16/16 \\
250 & -- & -- & -- & 250 & \cellcolor{accC}12/16 & \cellcolor{accA}16/16 & \cellcolor{accC}9/16 & 250 & \cellcolor{accC}12/16 & \cellcolor{accA}16/16 & \cellcolor{accA}16/16 \\
300 & -- & -- & -- & 300 & \cellcolor{accC}11/16 & \cellcolor{accA}16/16 & \cellcolor{accC}10/16 & 300 & \cellcolor{accC}10/16 & \cellcolor{accA}16/16 & \cellcolor{accA}16/16 \\
350 & -- & -- & -- & 350 & \cellcolor{accC}10/16 & \cellcolor{accA}16/16 & \cellcolor{accC}9/16 & 350 & \cellcolor{accC}12/16 & \cellcolor{accA}16/16 & \cellcolor{accA}16/16 \\
400 & -- & -- & -- & 400 & \cellcolor{accC}9/16 & \cellcolor{accA}16/16 & \cellcolor{accC}8/16 & 400 & \cellcolor{accC}11/16 & \cellcolor{accA}16/16 & \cellcolor{accA}16/16 \\
450 & -- & -- & -- & 450 & \cellcolor{accC}10/16 & \cellcolor{accA}16/16 & \cellcolor{accC}9/16 & 450 & \cellcolor{accB}14/16 & \cellcolor{accA}16/16 & \cellcolor{accA}16/16 \\
\cellcolor{selectedStep}\textbf{500} & \cellcolor{accA}\textbf{16/16} & \cellcolor{accD}\textbf{5/16} & \cellcolor{accE}\textbf{2/16} & \cellcolor{selectedStep}\textbf{500} & \cellcolor{accC}\textbf{8/16} & \cellcolor{accA}\textbf{16/16} & \cellcolor{accD}\textbf{7/16} & 500 & \cellcolor{accC}12/16 & \cellcolor{accA}16/16 & \cellcolor{accA}16/16 \\
550 & -- & -- & -- & 550 & \cellcolor{accC}10/16 & \cellcolor{accA}16/16 & \cellcolor{accC}8/16 & 550 & \cellcolor{accB}14/16 & \cellcolor{accA}16/16 & \cellcolor{accA}16/16 \\
600 & -- & -- & -- & 600 & \cellcolor{accC}10/16 & \cellcolor{accA}16/16 & \cellcolor{accC}8/16 & 600 & \cellcolor{accB}13/16 & \cellcolor{accA}16/16 & \cellcolor{accA}16/16 \\
650 & -- & -- & -- & 650 & \cellcolor{accC}11/16 & \cellcolor{accA}16/16 & \cellcolor{accC}9/16 & \cellcolor{selectedStep}\textbf{650} & \cellcolor{accC}\textbf{11/16} & \cellcolor{accA}\textbf{16/16} & \cellcolor{accA}\textbf{16/16} \\
700 & -- & -- & -- & 700 & \cellcolor{accC}8/16 & \cellcolor{accA}16/16 & \cellcolor{accC}12/16 & 700 & \cellcolor{accB}13/16 & \cellcolor{accA}16/16 & \cellcolor{accA}16/16 \\
\bottomrule
\end{tabular}
\end{adjustbox}
\end{table*}


\begin{table*}[!p]
\centering
\small
\setlength{\tabcolsep}{4.0pt}
\renewcommand{\arraystretch}{1.08}

\caption{
Reverse interference when re-training Q1 from the selected Q3@650 policy
on Qwen3-VL-8B.
The left block reports verifier-confirmed correct rollouts for Q1 and four
previously mastered variants from the Q2/Q3-related problem family.
The right block reports performance on DEV192 during Q4 training.
Cell colors follow the same absolute rollout-accuracy scale as
Tables~\ref{tab:2b_training_dynamics}--\ref{tab:8b_training_dynamics}.
Re-learning Q1 restores its direct-solving performance, but simultaneously
suppresses the previously mastered related family and degrades held-out
development performance.
}
\label{tab:q4_reverse_interference}

\begin{adjustbox}{max width=\textwidth}
\begin{tabular}{c|c|cccc|c|cccc}
\toprule

&
\multicolumn{6}{c|}{\textbf{Rollout Profile}}
&
\multicolumn{4}{c}{\textbf{Held-out Development Performance}}
\\

\cmidrule(lr){2-7}
\cmidrule(lr){8-11}

\textbf{Step}
&
\textbf{Q1}
&
\textbf{Q3/base}
&
\textbf{420/xy}
&
\textbf{Q2/y}
&
\textbf{420/z}
&
\textbf{Family Total}
&
\textbf{MATH96}
&
\textbf{WE-MATH96}
&
\textbf{DEV192}
&
$\boldsymbol{\Delta}$ \textbf{DEV}
\\
\midrule

\textbf{0 (Q3@650)}
&
\cellcolor{accC}\textbf{11/16}
&
\cellcolor{accA}\textbf{16/16}
&
\cellcolor{accA}\textbf{16/16}
&
\cellcolor{accA}\textbf{16/16}
&
\cellcolor{accA}\textbf{16/16}
&
\textbf{64/64}
&
86.4583
&
69.7900
&
\textbf{78.1242}
&
0
\\

25
&
\cellcolor{accC}10/16
&
\cellcolor{accD}6/16
&
\cellcolor{accD}4/16
&
\cellcolor{accD}6/16
&
\cellcolor{accD}6/16
&
22/64
&
81.2500
&
66.6700
&
73.9600
&
-4.1642
\\

50
&
\cellcolor{accC}11/16
&
\cellcolor{accD}7/16
&
\cellcolor{accE}2/16
&
\cellcolor{accD}5/16
&
\cellcolor{accD}5/16
&
19/64
&
84.3750
&
65.6200
&
74.9975
&
-3.1267
\\

75
&
\cellcolor{accC}11/16
&
\cellcolor{accC}9/16
&
\cellcolor{accD}4/16
&
\cellcolor{accD}5/16
&
\cellcolor{accD}7/16
&
25/64
&
84.3750
&
67.7100
&
76.0425
&
-2.0817
\\

100
&
\cellcolor{accC}12/16
&
\cellcolor{accD}6/16
&
\cellcolor{accE}2/16
&
\cellcolor{accD}5/16
&
\cellcolor{accD}4/16
&
\textbf{17/64}
&
83.3333
&
63.5400
&
73.4367
&
-4.6875
\\

125
&
\cellcolor{accA}\textbf{16/16}
&
\cellcolor{accC}10/16
&
\cellcolor{accE}3/16
&
\cellcolor{accD}6/16
&
\cellcolor{accD}5/16
&
24/64
&
81.2500
&
62.5000
&
\textbf{71.8750}
&
\textbf{-6.2492}
\\

150
&
\cellcolor{accB}14/16
&
\cellcolor{accC}10/16
&
\cellcolor{accE}2/16
&
\cellcolor{accD}5/16
&
\cellcolor{accD}6/16
&
23/64
&
83.3333
&
64.5800
&
73.9567
&
-4.1675
\\

175
&
\cellcolor{accB}14/16
&
\cellcolor{accD}7/16
&
\cellcolor{accE}3/16
&
\cellcolor{accD}6/16
&
\cellcolor{accD}4/16
&
20/64
&
84.3750
&
65.6200
&
74.9975
&
-3.1267
\\

200
&
\cellcolor{accA}\textbf{16/16}
&
\cellcolor{accC}9/16
&
\cellcolor{accD}4/16
&
\cellcolor{accD}5/16
&
\cellcolor{accD}5/16
&
\textbf{23/64}
&
82.2917
&
67.7100
&
75.0008
&
-3.1233
\\

\bottomrule
\end{tabular}
\end{adjustbox}

\vspace{1.5mm}
\footnotesize
\textbf{Full-test shift at Q4@200:}
MATH500 $86.80 \rightarrow 84.80$ ($-2.00$);
WE-MATH testmini $73.51 \rightarrow 72.13$ ($-1.38$);
Full Avg. $80.155 \rightarrow 78.465$ ($-1.690$).

\end{table*}

\begin{table*}[t]
\centering
\caption{Rollout-budget accounting for the scheduling baselines. ``Prompt slots'' denotes the total number of training prompts processed before rollout expansion. Policy-Adaptive Sequential RLVR and Full-Batch GRPO are exactly matched at 33,600 rollout responses. Fixed-Sequential follows the original 700-step protocol and uses 11,200 rollouts.}
\label{tab:rollout_budget}
\begin{adjustbox}{max width=0.95\linewidth}
\begin{tabular}{llccccc}
\specialrule{.16em}{0pt}{.65ex}
\textbf{Backbone} & \textbf{Method} & \textbf{Training Schedule} & \textbf{Batch} & \textbf{Rollout $n$} & \textbf{Prompt Slots} & \textbf{Total Rollouts} \\
\specialrule{.05em}{.4ex}{.65ex}
2B & Policy-Adaptive Sequential & $3\times700$ & 1 & 16 & 2,100 & \textbf{33,600} \\
2B & Fixed-Sequential & 700 & 1 & 16 & 700 & 11,200 \\
2B & Full-Batch GRPO & $30\times69+30$ & $69\rightarrow30$ & 16 & 2,100 & \textbf{33,600} \\
\midrule
4B & Policy-Adaptive Sequential & $3\times700$ & 1 & 16 & 2,100 & \textbf{33,600} \\
4B & Fixed-Sequential & 700 & 1 & 16 & 700 & 11,200 \\
4B & Full-Batch GRPO & $65\times32+20$ & $32\rightarrow20$ & 16 & 2,100 & \textbf{33,600} \\
\midrule
8B & Policy-Adaptive Sequential & $3\times700$ & 1 & 16 & 2,100 & \textbf{33,600} \\
8B & Fixed-Sequential & 700 & 1 & 16 & 700 & 11,200 \\
8B & Full-Batch GRPO & $75\times28$ & 28 & 16 & 2,100 & \textbf{33,600} \\
\specialrule{.16em}{.65ex}{0pt}
\end{tabular}
\end{adjustbox}
\end{table*}

\end{document}